\pdfoutput=1

\documentclass[x11names]{article}

\usepackage[preprint]{acl}

\usepackage{times}
\usepackage{latexsym}

\usepackage[T1]{fontenc}

\usepackage[utf8]{inputenc}

\usepackage{microtype}

\usepackage{inconsolata}

\usepackage{graphicx}

\usepackage{booktabs} 
\usepackage{multirow} 
\usepackage{multicol} 
\usepackage{amsmath}
\usepackage{enumitem} 
\usepackage{todonotes}
\usepackage{hyperref} 

\title{Large Language Models as a Tool for Mining Object Knowledge}

\author{Hannah YoungEun An \\
University of Rochester\\
\texttt{yan2@cs.rochester.edu} \\\And
Lenhart K. Schubert \\
University of Rochester \\
\texttt{schubert@cs.rochester.edu} \\}

\begin{document}
\maketitle

\begin{abstract}
Commonsense knowledge is essential for machines to reason about the world. Large language models (LLMs) have demonstrated their ability to perform almost human-like text generation. Despite this success, they fall short as trustworthy intelligent systems, due to the opacity of the basis for their answers and a tendency to confabulate facts when questioned about obscure entities or technical domains. We hypothesize, however, that their \textit{general} knowledge about objects in the everyday world is largely sound. Based on that hypothesis, this paper investigates LLMs' ability to formulate explicit knowledge about common physical artifacts, focusing on their parts and materials. Our work distinguishes between the substances that comprise an entire object and those that constitute its parts---a previously underexplored distinction in knowledge base construction. Using few-shot with five in-context examples and zero-shot multi-step prompting, we produce a repository of data on the parts and materials of about 2,300 objects and their subtypes. Our evaluation demonstrates LLMs' coverage and soundness in extracting knowledge. This contribution to knowledge mining should prove useful to AI research on reasoning about object structure and composition and serve as an explicit knowledge source (analogous to knowledge graphs) for LLMs performing multi-hop question answering.
\end{abstract}

\section{Introduction}
Recently, large language models (LLMs) have gained considerable attention in the natural language processing (NLP) community for generating almost human-like text and generalizing their knowledge for NLP downstream tasks \citep{bhavya-etal-2022-analogy, bubeck-etal-2023-sparks, moghaddam-honey-2023-boosting}. However, one reason that LLMs fall short of artificial general intelligence is that the knowledge underlying their responses is entirely implicit, aside from the ongoing dialogue contents they directly reference. As a result, they are apt to provide flawed justifications for both correct and incorrect conclusions involving step-by-step commonsense reasoning, without allowing users ``patch'' faulty assumptions or inference steps. Additionally, they tend to hallucinate facts, especially regarding named entities sparsely represented on the web.

In this work, we study the feasibility of generating explicit general knowledge, drawing on language models' implicit knowledge and language understanding ability, specifically for knowledge about the part structure and material composition of physical objects. 
Such knowledge, common even to young children, enables explainable reasoning and decision-making in combination with other commonsense knowledge, e.g., in considering whether scissors can be used to shorten a shoestring or a screw,
or if a basketball or a wine bottle can be pushed off a patio table safely to the flagstone surface beneath.
While LLMs may often answer adequately in such cases, when they don't, they cannot simply be told how to correct their knowledge; unlike people, they learn by abstraction from massive data rather than by simply being told.

While there has been work on mining or crowdsourcing semantic features that characterize human conceptual knowledge, no large-scale existing work makes a distinction between the substances that comprise an entire object and the substances that comprise an object's parts. For example, a violin is mostly made of wood; however, more precisely, the body, fingerboard, pegs, and many other components of are made of wood while the strings are made from non-wooden materials, such as catgut, nylon, and steel. Our approach mines part and material knowledge from LLMs and expresses it in an explicit form. This knowledge can then be integrated into the knowledge base of AI systems, either supplementing or replacing black-box models and potentially enhancing explainability. From the resulting resource, which encompasses the data of 2,314 physical objects, we evaluated a subset of these items for coverage and quality through cross-dataset comparisons and intrinsic assessments.
The results indicate that the majority of the extracted knowledge aligns with human understanding, but depending on the prompting method used, the knowledge can sometimes be overly simplified or excessively specific, exceeding typical layman knowledge. All data and source code are publicly available to support further research\footnote{\href{https://anonymous.4open.science/r/composition-miner}{\texttt{https://anonymous.4open.science/r/composition\linebreak-miner}}}.

\section{Related Work}

\subsection{Pattern-based extraction}
The pattern-based approach in knowledge acquisition identifies relationships in text by matching predefined lexical patterns. While it offers consistent results, it struggles with complex data patterns and implicit knowledge that is typically not documented. Early efforts, like \citet{berland-charniak-1999-finding} and \citet{poesio-etal-2002-acquiring}, relied on hand-built patterns to detect part-whole relationships. Later, search by \citet{girju-etal-2006-automatic} and \citet{van-hage-etal-2006-method} automated the process using web data, refining it further with classifiers to determine meronymy relationships. \citeauthor{girju-etal-2006-automatic} used a two-way classifier, while \citet{poesio-almuhareb-2005-identifying} employed a multi-way classifier to categorize detected attributes. More recent advancements, such as \citet{tesfaye-zock-2012-automatic}, used vector similarity to cluster co-occurring nouns, enhancing both accuracy and coverage.

\begin{figure*}[!ht]





    \centering
    \includegraphics[width=0.97\textwidth]{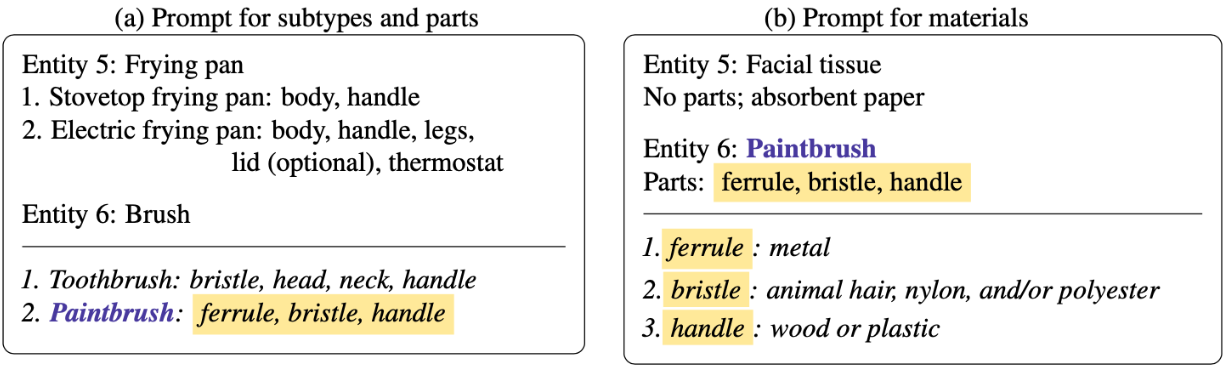}
    \caption{A simplified example of our few-shot prompting approach. In both (a) and (b), the input to the LLM appears above the line, and the output below. (a) illustrates subtype classification and parts identification, identifying a paintbrush as a subtype of a brush. (b) focuses on material identification, listing the materials associated with paintbrush's parts extracted from (a).}
    \label{fig:few-shot}
\end{figure*}

\subsection{Human annotation}
Human annotation often involves significant expenses and time due to the need for detailed and accurate annotations. When data is collected through crowdsourcing, ensuring quality control becomes challenging, especially when complex semantic relationships need to be precisely captured.

Several knowledge resources integrate crowd-sourced and expert annotations to address these challenges. ConceptNet \citep{speer-etal-2017-conceptnet} incorporates both methods to build a commonsense knowledge graph, linking concepts with relations like \verb|MadeOf| and \verb|PartOf|. However, the annotation schema can lead to ambiguity; \verb|MadeOf| describes an object as a whole, while \verb|PartOf| identifies components without specifying their materials, even when the parts are made of different materials from other parts. For example, the tuples (\texttt{bicycle}, \texttt{MadeOf}, \texttt{metal}) and (\texttt{bicycle seat}, \texttt{PartOf}, \texttt{bicycle}) does not specify a bicycle saddle, which is a part of a bicycle seat, can be made of leather.
WordNet \citep{miller-1995-wordnet} provides an expert-curated lexical database, where human annotators organize words into semantic hierarchies and mark part-whole relations (meronymy). Its concise definitions sometimes encode part and material information, as in the example of `felt-tip pen,' whose definition states the writing tip is made of felt.
Similarly, the ParRoT dataset, developed by \citet{gu-etal-2023-language}, offers a fine-grained, human-annotated collection of part lists for 100 everyday objects. This dataset goes beyond major components, detailing sub-parts (e.g., the reflective glass and spring in a flashlight) to provide a comprehensive model of objects, particularly in terms of their structural and functional relationships.
Semantic feature datasets, such as McRae norms \citep{mcrae-etal-2005-semantic} and CSLB concept property norms \citep{devereux-etal-2014-centre}, rely on human expertise to generate reliable annotations. These datasets include features like part-whole relations and material composition such as \texttt{has\_a} and \texttt{made\_of}. The McRae norms distinguish between \texttt{made\_of}, used for substances, and \texttt{made\_from}, used for origins (e.g., prune made from plums). Additionally, they differentiate between essential components (e.g., an engine or a door) and non-essential parts (e.g., a bed's comforter) or functional aspects (e.g., an elevator's capacity), assigning them distinct labels.


\subsection{Use of LLMs}
LLMs excel at capturing complex patterns and implicit world knowledge from their pre-training, enabling them to generate and retrieve comprehensive information with high linguistic fluency. However, challenges remain, including issues with data interpretability and the risk of generating factually inaccurate content. Several studies demonstrate both the potential and limitations of these models in generating and structuring such knowledge.

TransOMCS \citep{zhang-etal-2020-transomcs} builds a large commonsense graph by extracting knowledge from linguistic data but struggles with misinterpretations, such as (\texttt{cement}, \texttt{MadeOf}, \texttt{consist}) and (\texttt{dog}, \texttt{PartOf}, \texttt{walker}), illustrating both scalability and the risks of generating incorrect data.
$\text{ATOMIC}_{20}^{20}$ \citep{hwang-etal-2020-cometatomic} combines crowdsourcing with LLMs, creating 1.33 million tuples. However, it introduces imprecision by merging both part and material information under the \texttt{MadeUpOf} relation. Similarly, part-related knowledge is sometimes merged with other properties under \texttt{HasProperty}, e.g., (\texttt{bicycle}, \texttt{HasProperty}, \texttt{two wheels}).
ASCENT++ \citep{nguyen-etal-2022-refined} refines knowledge extraction through clustering, yet still produces odd outputs, such as (\texttt{crane}, \texttt{MadeOf}, \texttt{many different colors}) and (\texttt{desk}, \texttt{HasA}, \texttt{multilingual staff}), reflecting persistent noise.
Meanwhile, \citet{hansen-hebart-2022-method} employ GPT-3 to generate semantic features for 1,854 object concepts, producing detailed outputs (e.g., `scissors have two blades, handles, and a pivot point'). However, material descriptions about parts remain imperfect (e.g., `scissors are made of metal', missing mention of plastic handles), and occasional incoherent responses appear (e.g., `Wineglass is a glass, it is clear...What are the properties of a bat? It is used in sport').
These examples demonstrate LLMs' potential for structuring knowledge while highlighting ongoing issues with precision and noisy outputs.

\section{Method}
In this section, we describe the methodologies used to identify subtypes, parts, and materials of various entities. We employ few-shot in-context learning and zero-shot multi-step learning, both leveraging GPT-4 Turbo (\texttt{gpt-4-1106-preview}) \citep{openai-2023-gpt4}. The process begins with constructing a list of common physical entities, followed by few-shot and zero-shot learning independently, to generate separate datasets. These datasets allow us to compare their outcomes and ensure wide coverage in object classification. The prompts used for both methods are provided in Appendices \ref{sec:fewshot-prompt} and \ref{sec:zeroshot-prompt}.

\subsection{Common physical objects}
\label{sec:common-objects-short}
To gather commonsense knowledge about the material composition of objects, we focus on artifacts since their part structure, and especially their material composition, is generally clearer than those of natural objects. Starting with entities from Wikidata \citep{vrandecic-2014-wikidata}, we select those classified as artificial physical objects or structures. Abstract entities are removed through subclass filtering and keyword matching. We then further refine by retaining only entries that exist in WordNet and have corresponding Wikipedia links. Finally, GPT-4 \citep{openai-2023-gpt4} filters the list to include common, physical, standalone objects, resulting in a list of 2,314 entries. Full details of the filtering process are provided in Appendix \ref{sec:common-objects-long}.

\definecolor{babyblueeyes}{rgb}{0.63, 0.79, 0.95}
\definecolor{mossgreen}{rgb}{0.68, 0.87, 0.68}
\usetikzlibrary{arrows.meta}

\tikzset{
  treenode/.style = {shape=rectangle, rounded corners,
                     draw, align=center, font=\ttfamily\small},
  root/.style     = {treenode, fill=red!30},
  dec/.style      = {treenode, fill=babyblueeyes},
  destin/.style   = {treenode, fill=mossgreen},
  arrow/.style={->,>=stealth, arrows={-Latex[angle=40:2mm]}}
}

\begin{figure*}[!ht]
\centering
\includegraphics[width=0.99\textwidth]{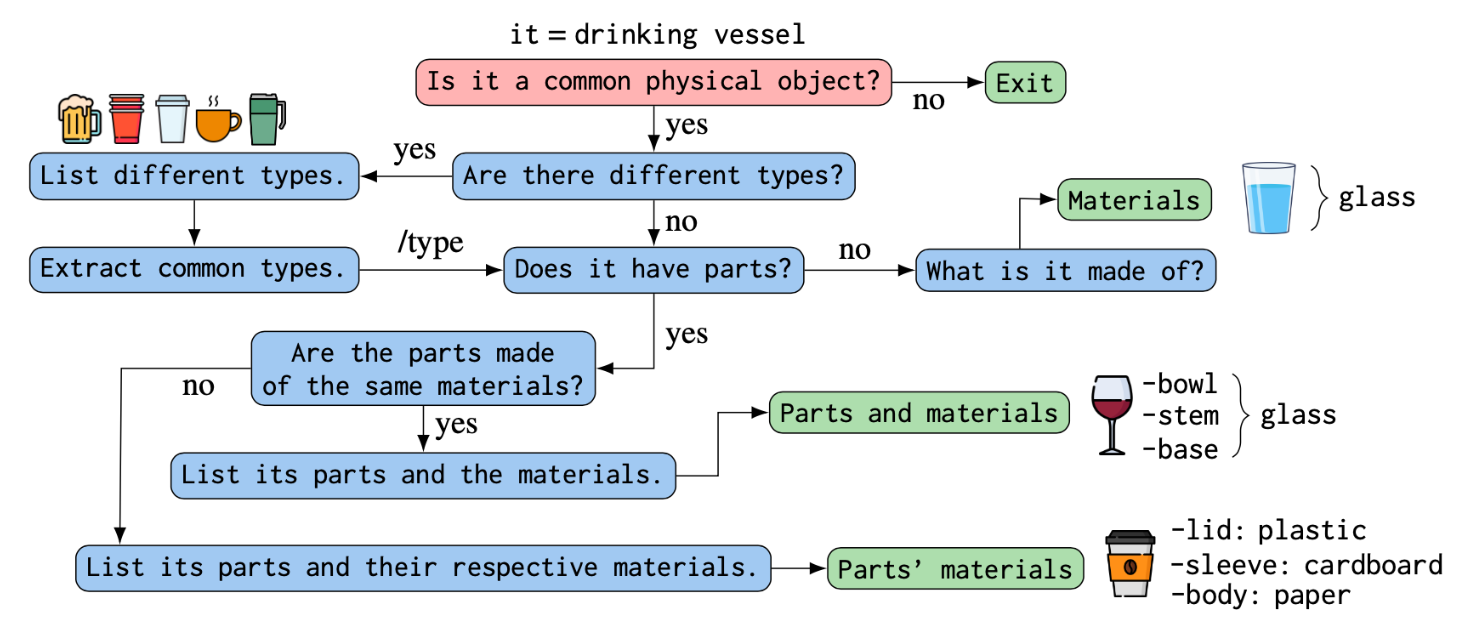}
\caption{Illustration of our zero-shot prompting using a multi-step classification algorithm to acquire subtype, part, and material information from GPT-4 Turbo.}
\label{fig:multistep-algo}
\end{figure*}

\subsection{Few-shot in-context learning}
We employ a few-shot learning method, using five in-context examples, to classify subtypes of entities based on their essential parts and identify those parts and the materials commonly used in them. This approach is divided into two stages. In the first stage, the model is prompted to enumerate common subtypes and their constituent parts for each entity. Subtype classification is strictly based on the presence or absence of essential, unique parts, explicitly excluding non-essential variations in size, shape, material, or function. If an entity has no subtypes, the model is instructed to specify this and list only the parts. In the second stage, after identifying subtypes and their parts, we extract typical materials used for these parts, excluding any that are primarily used for joining, stitching, or finishing. The model is guided to use specific conjunctions (`and,' `or,' `and/or') to reflect exclusivity, common combinations, or required co-occurrence of materials. A simplified example of the prompt with one in-context demonstration and its corresponding output is shown in Figure \ref{fig:few-shot}.

\subsection{Multi-step prompting}
We also employ a zero-shot prompting that is broken down into a sequence of sub-questions to identify and categorize subtypes, parts, and materials. Although each sub-question operates under a zero-shot framework, the multi-step sequence is designed to guide the LLM through complex classification tasks in a structured and incremental manner. This approach ensures that the model handles multifaceted object classifications step-by-step. Figure \ref{fig:multistep-algo} provides an overview of the zero-shot multi-step classification algorithm, illustrating how each step builds on the previous one to refine the categorization process. The process begins with the list of common physical objects identified in Section \ref{sec:common-objects-short}, followed by steps to classify their subtypes and constituent elements, as detailed below.
\vspace{-2mm}
\begin{enumerate}[leftmargin=*] 
\itemsep-0.5mm
\item \textbf{Are there different types?} This first prompt asks whether the entity can be divided into distinct subtypes, ensuring that only objects with meaningful variation in their essential parts proceed through further classification steps.
\item \textbf{List different types.} If the entity is determined to have subtypes, the next step involves prompting the model to generate a comprehensive list of those subtypes. The prompt emphasizes focusing on physically distinct subtypes, filtering out superficial design variations or synonyms.
\item \textbf{Extract common types.} Once subtypes are identified, this prompt asks how recognizable each subtype is to the general public. This step narrows the focus to widely recognized objects.
\item \textbf{Does it have parts}? For each identified subtype (or for the original entity if no subtypes are found), this prompt asks whether it has multiple, clearly distinct parts.
\item \textbf{Are the parts made of the same materials?} If the entity or subtype has parts, it checks whether these parts are composed of the same materials.
\item \textbf{What is it made of?} If the entity/subtype does not distinct parts, the model is prompted to list the materials that are typically used in it.
\item \textbf{List its parts and the materials.} This prompt gathers a detailed list of the object's parts and their material composition for the entire object.
\item \textbf{List its parts and their respective materials.} This prompt provides a granular view by identifying the object's parts and the specific materials used for each individual part.
\end{enumerate}
\vspace{-3mm}

\begin{table*}[!ht]
  \centering
    \begin{tabular}{cccllll}
    \hline
    \multirow{2}{*}{\textbf{Category}} & \multirow{2}{*}{\textbf{Our data}} & \multicolumn{5}{c}{\textbf{Reference data}}\\
    & & ParRoT & CSLB & McRae & WordNet & ConceptNet\\
    \hline
    \multirow{2}{*}{Part} & Few-shot & 58.54 (164) & 65.57 (106) & 77.08 (24) & 75.47 (53) & 78.33 (30) \\
    & Zero-shot & 91.67 (174) & 95.87 (109) & 96.00 (25) & 87.74 (53) & 95.00 (30)\\
    \hline
    \multirow{2}{*}{Material} & Few-shot & - & 88.81 (67) & 98.28 (29) & 76.67 (15) & 96.15 (13)\\
    & Zero-shot & - & 97.14 (70) & 98.33 (30) & 96.67 (15) & 96.15 (13)\\

  \end{tabular}
  \caption{Recall values for external datasets calculated based on part and material availability in our dataset. Items receive full credit (1), half credit (0.5), or no credit (0) based on their presence. Recall is calculated as the total score divided by the total number of parts and materials from reference datasets, which are listed in parentheses next to each recall value. A dash (`-') indicates that the ParRoT dataset did not contain material information.}
  \label{table:external-recall}
\end{table*}


\section{Results \& Analysis}
The acquired datasets are hierarchically organized into entities (Wikipedia entries), subtypes, and subsubtypes, with parts and materials listed at the lowest level. The few-shot data includes over 6,000 items, while zero-shot data covers nearly 27,300 items. Both datasets feature thousands of subtypes, though the zero-shot data is more extensive. Most items list multiple parts and materials, with zero-shot items averaging around eight parts and two materials, while few-shot items tend to have slightly less. For additional specifics, refer to Appendix \ref{sec:overview-distribution}.

\subsection{Recall based on external datasets}
To evaluate coverage of our dataset, we calculate cross-dataset recall by comparing items from five reference datasets---ParRoT, CSLB, McRae, WordNet, and ConceptNet---to our own. For ConceptNet, we exclude entries imported from WordNet to avoid double counting.
A random selection of 20 entities per dataset is used, replacing items without matches in our dataset or lacking relevant data. For example, some entities from WordNet and McRae (e.g., socks and diving suit) are replaced due to missing both part and material data, while 37 ConceptNet entities (e.g., helmet and microwave) are similarly excluded. Duplicate items (e.g., power supply cord and power cord), design- or shape-related features (e.g., open top of a bucket), and items that are not physically attached (e.g., airplane pilot) are also removed. ConceptNet's data quality is inconsistent, with problematic entries like \textit{no lead} as part of a pencil and \textit{accent on second syllable} as part of an umbrella. In the case of WordNet, part and material information embedded within glosses is treated as relevant information.

To measure cross-dataset recall, we score each reference item based on its presence in our dataset: full credit (1 point), half credit (0.5 points), and no credit (0 points). Recall is calculated as the total score divided by the total number of reference items. Full credit is assigned when 1) a matching item is found (e.g., \textit{lens of spectacles} from WordNet matches \textit{lenses of glasses}), 2) the reference item is a substance and our dataset contains a suitable container for it (e.g., \textit{propellant of a fire extinguisher} from ParRoT matches \textit{pressure cartridge}), or 3) a reference item, which is a supertype of another reference item, is satisfied by an item in our dataset not mentioned in the reference dataset (e.g., when \textit{metal, aluminium, and light metal} is listed, we interpret \textit{light metal} as ``aluminium and some other metals'', so \textit{light metal} matches \textit{titanium}). Half credit is given when 1) the item in our dataset is inclusive of the reference item (e.g., \textit{seats of an airplane} from CSLB matches \textit{cabin of an airplane}; \textit{straw used in a hut} from McRae matches \textit{thatch}) or 2) our item represents a more general type of the reference item (e.g., \textit{side carry handle of a suitcase} from ParRoT matches \textit{handle}; \textit{rubberized fabric of a raincoat} from WordNet matches \textit{waterproof fabric}).

In Table \ref{table:external-recall}, we report the cross-dataset recall values along with the total number of reference items in parentheses. The recall for zero-shot is consistently higher than few-shot in both part and material data categories, indicating that the zero-shot approach captures more parts and materials that are available in the reference datasets. Certain datasets like ParRoT and CSLB show a significant improvement from few-shot to zero-shot, especially in the part data. These datasets contain more complex items, which benefit the zero-shot approach. For example, ParRoT specifies \textit{apron, cross rail, top rail}, and \textit{stile} for a chair, in addition to \textit{back, legs}, and \textit{seat}, while our few-shot results tend to include only broader categories like \textit{legs (or base), seat, backrest}, and \textit{armrests}, lacking the more detailed part distinctions.


\subsection{Intrinsic evaluation}
\label{sec:intrinsic-evaluation}
To evaluate the quality of our dataset, we conduct intrinsic evaluations using three metrics: precision, recall, and distinctive feature evaluation. We sample 120 entities from the list of common physical objects obtained in Section \ref{sec:common-objects-short} and apply a filtering process to ensure that the number of items evaluated does not exceed 1,000 per category, with a few exceptions (details in Appendix \ref{sec:intrinsic-evaluation-breakdown}). The intrinsic evaluation tasks are conducted on Amazon Mechanical Turk (MTurk), where workers answer multiple-choice questions, with responses collected from three different workers for each question.
Sample questions and a detailed breakdown of the collected responses are provided in Appendix \ref{sec:mturk-samples} and \ref{sec:experiment-results}.

\paragraph{Precision}
\begin{table}
  \centering
  \begin{tabular}{c@{\hskip0.1in}c@{\hskip0.1in}cc@{\hskip0.1in}c}
    \hline
    \textbf{Data} & \textbf{Category} & \textbf{Subtype} & \textbf{Part} & \textbf{Material}\\
    \hline
    & Likely & 84.96 & 84.79 & 88.27\\
    Few & Unlikely & 2.09 & 6.20 & 3.42\\
    shot & Unable & 12.02 & 8.90 & 8.30\\
    & Uncertain & 0.93 & 0.11 & 0.00 \\
    \hline
    & Likely & 91.63 & 79.90 & 88.77\\
    Zero & Unlikely & 3.32 & 9.37 & 3.53\\
    shot & Unable & 4.64 & 10.47 & 7.67\\
    & Uncertain & 0.41 & 0.27 & 0.03\\
  \end{tabular}
  \caption{Precision for the subtypes, parts, and materials in the few-shot and zero-shot data. The table shows the percentage of responses marked as \textit{Likely}, \textit{Unlikely}, \textit{Unable (to answer)}, and \textit{Uncertain} for each category.}
  \label{table:internal-precision}
\end{table}

\begin{table}[!ht]
  \centering
  \begin{tabular}{llc}
    \hline
    \textbf{Unlikely answer detail} & \textbf{Few} & \textbf{Zero}\\
    \hline
    Item does not have parts & 0.00 & 1.07\\
    Part is rather a feature & 47.67 & 44.13\\
    Part is considered a material & 13.95 & 6.41\\
    Not an essential part & 30.81 & 32.03\\
    Irrelevant part & 4.65 & 11.03\\
    Other reasons & 2.91 & 5.34\\
  \end{tabular}
  \caption{Distribution of reasons for \textit{Unlikely} answers in precision evaluation of `part' items. The table presents the proportion of responses for each reason across the few-shot and zero-shot data.}
  \label{table:internal-precision-part-breakdown}
\end{table}

\begin{table*}
  \centering
  \begin{tabular}{lcccccc}
    \hline
    \multirow{2}{*}{\textbf{Response category}} & \multicolumn{2}{@{\hskip0.2em}c}{\textbf{Subtype}} & \multicolumn{2}{@{\hskip-0.8em}c}{\textbf{Part}} & \multicolumn{2}{@{}c}{\textbf{Material}}\\
    & Few & Zero & Few & Zero & Few & Zero\\
    \hline
    Likely with no issues & 49.54 & 44.86 & 58.17 & 74.53 & 71.43 & 73.67\\
    Likely, even though some items overlap & 5.56 & 13.58 & 0.89 & 2.57 & 0.62 & 1.13\\
    Unlikely, one major item is missing & 19.44 & 20.99 & 21.12 & 8.17 & 15.65 & 12.90\\
    Unlikely, two major items are missing & 5.56 & 9.47 & 5.02 & 2.10 & 3.19 & 2.90\\
    Unlikely, three or more major items are missing & 8.80 & 6.17 & 2.67 & 1.57 & 1.23 & 0.70\\
    Unlikely for other reasons & 0.93 & 0.41 & 0.08 & 0.40 & 0.38 & 0.13\\
    Unable to answer & 10.19 & 4.53 & 11.97 & 10.67 & 7.42 & 8.50\\
    Uncertain & 0.00 & 0.00 & 0.08 & 0.00 & 0.08 & 0.07\\
  \end{tabular}
  \caption{Recall for the lists of subtypes, parts, and materials in the few-shot and zero-shot data.}
  \label{table:internal-recall}
\end{table*}

We evaluate whether an item is likely to be a subtype, part, or material of a given item. Workers are asked to categorize items according to multiple-choice options, including \textit{Likely}, \textit{Unlikely}, \textit{Unable (to answer)}, and \textit{Uncertain}. The precision results (Table \ref{table:internal-precision})  show that most items were marked as \textit{Likely} across all three categories (Subtype, Part, and Material). In particular, the zero-shot data received a higher portion of \textit{Likely} responses than the few-shot data, especially for subtypes.
In the few-shot data, a notable portion of responses for subtypes were marked as \textit{Unable}, which may indicate that workers were unfamiliar with specific items or found the subtype names unclear or confusing.
For the Part category, there is a higher percentage of \textit{Unlikely} and \textit{Unable} responses, compared to the other categories. This suggests that identifying essential parts poses more challenges for both LLMs and human evaluators, possibly due to the complexity involved in reasoning about parts. Interestingly, all categories had a very low percentage of responses marked as \textit{Uncertain} (less than 1\%), for both few-shot and zero-shot data. This suggests that, while there is some unfamiliarity, workers rarely expressed outright uncertainty about their evaluations.

To better understand the reasoning behind the significant portion of \textit{Unlikely} responses for the part items, we further analyze these \textit{Unlikely} responses. Table \ref{table:internal-precision-part-breakdown} provides a detailed breakdown of the specific reasons why certain parts were deemed unlikely, and the respective proportions for each reason. The most dominant reason that the parts are labeled with \textit{Unlikely} in both few-shot and zero-shot data (47.67\% and 44.13\%, respectively), was that the workers identified the parts listed in the dataset as a feature rather than a part. There are also a substantial portion (30.81\% and 32.03\%) of the parts identified as \textit{Not an essential part}; workers judged many of the parts listed by the LLM as peripheral to the item's existence, i.e., the part is often shown/used with the item, but it is either not essential, or not included with or attached to the item. The same analysis for material items, which had relatively few \textit{Unlikely} responses, is provided in Appendix \ref{sec:internal-precision-material-breakdown}.

\begin{table}
  \centering
  \begin{tabular}{llc}
    \hline
    \textbf{Response category} & \textbf{Few} & \textbf{Zero}\\
    \hline
    Likely & 31.18 & 20.64\\
    Unlikely, part is present in both & 48.90 & 58.04\\
    Unlikely, part is absent in both & 1.01 & 2.30\\

    Unlikely for other reasons & 0.17 & 0.40\\
    Unable to answer & 16.75 & 17.13\\
    Uncertain & 1.98 & 1.49\\
  \end{tabular}
  \caption{Distribution of responses in evaluating distinctive feature significance, based on reasons for identifying a main distinguishing part between two subtypes.}
  \label{table:distinctiveness-evaluation}
\end{table}

\paragraph{Recall}
In a manner similar to the evaluation of precision, we assess whether the provided lists for subtypes, parts, or materials of an item are complete. The summarized results of the recall evaluation are presented in Table \ref{table:internal-recall}. Overall, LLM-generated subtype lists appear to be perceived as less complete compared to the parts or materials lists. This is particularly evident from the higher percentages in the \textit{one major item is missing} and \textit{two or more major items are missing} categories for subtypes, suggesting that workers often found the subtype lists to be insufficient.

For parts, the zero-shot method performed notably better, with a 74.53\% likelihood that the list was considered complete with no issues, compared to only 58.17\% in the few-shot setting. The few-shot method also showed a more frequent pattern of missing one, two, or even three or more major parts. This trend stands in contrast to the results for subtypes, where fewer items were missing in the few-shot condition than in the zero-shot condition. This difference suggests that the method used can have a substantial impact depending on the type of information (subtypes vs. parts) generated by the LLM.
Interestingly, material lists seem to perform the best across both few- and zero-shot settings, with fewer concerns about missing or overlapping items. This outcome makes sense, as the number of materials associated with an item or part tends to be smaller, making it easier to ensure the list is complete.

\begin{table*}[!ht]
  \centering
  \includegraphics[width=0.97\textwidth]{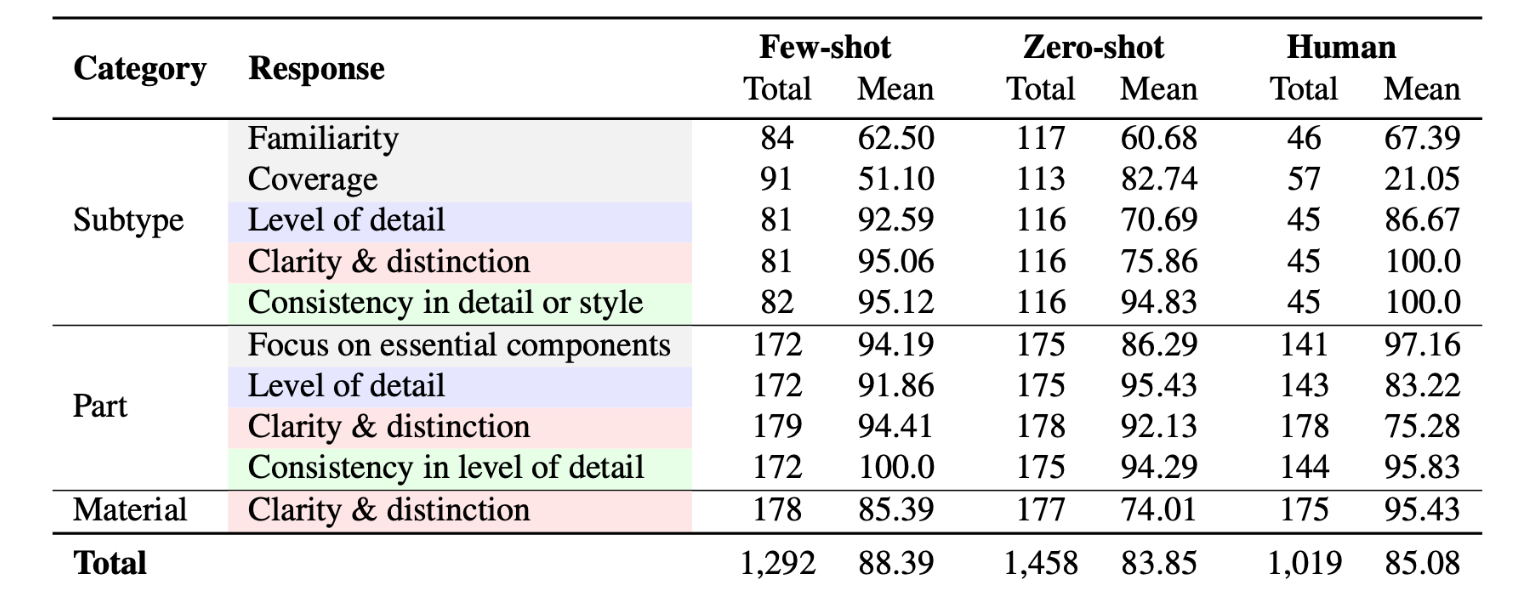}
  \caption{Comparison of three datasets (few-shot, zero-shot, and human-annotated data) across multiple response categories. The `Total' column represents the total number of responses recorded for each category, while the `Mean' column reflects the average score, calculated as the ration of the aggregate score to the total number of responses. The scores are derived based on predefined rating criteria. Details on the collected responses are in Appendix \ref{sec:dataset-comparison-breakdown}.}
  \label{table:dataset-comparison}
\end{table*}

\paragraph{Distinctive feature significance} 
This metric captures whether a specific part serves as a key distinguishing factor between two subtypes. Workers are presented with one subtype, a particular part that it has, and another subtype within the same supertype that lacks this part. Table \ref{table:distinctiveness-evaluation} summarizes the evaluation results, with a signification proportion of responses falling under \textit{Unlikely because the part is present in both} (48.90\% for few-shot and 58.04\% for zero-shot data). This indicates that the LLM often struggled to recognize that many of the parts it generated could apply to multiple subtypes, or at least failed to present the data with a clear distinction between which parts were unique to a subtype and which were shared. In the zero-shot setting, this is understandable since each question stands alone without prior context. However, the few-shot setting, which does contain comparative context in the response itself, still showed limitations in the model's ability to structurally distinguish unique or overlapping parts across subtypes.

\subsection{Dataset-comparison Evaluation}
\paragraph{Human-annotated data creation}
We also evaluate the quality of our dataset using comparisons with human annotations.
For this, we create a human-annotated dataset to capture the detailed part structures and material compositions of 120 entities. These entities are randomly sampled from the object list provided in Section \ref{sec:common-objects-short}, ensuring no overlap with the 120 entities used for intrinsic evaluation in the previous section. The annotation process followed a structured approach, conducted by in-house lab members and the authors, guided by a set of criteria (in Appendix \ref{sec:human-annotation-criteria}).
Annotators relied on Wikipedia as the primary reference source but were not limited to it; they were also allowed to consult external sources, such as online searches, when necessary.
Subtypes were annotated up to two levels---subtypes and subsubtypes---when relevant, ensuring consistency with the depth used in few-shot and zero-shot methods. Subtypes were included only when they reflected meaningful differences in part structure or material composition. For entries with existing Wikipedia pages, we strictly adhered to Wikipedia's naming conventions. Much like few-shot and zero-shot methods, our dataset captures detailed part structures, including optional components, and appropriately uses conjunctions like `and/or' to reflect material combinations\footnote{Annotators were allowed to use ``entity'' instead of a material to indicate that the part is composed of further parts}. We also annotated instances where uniform materials spanned multiple parts of an entity.

\paragraph{Comparison with human-annotated data}
To assess the quality and the effectiveness of the few-shot and zero-shot extraction methods, we designed an evaluation process that compares their results with human-annotated data. The evaluation was also conducted on MTurk, with three different workers providing responses for each question. We evaluated 60 entities, each appearing across three datasets, resulting in a total of 180 classifications. To ensure fairness, none of these entities were included in the few-shot prompts. For consistency, the same worker evaluated all questions related to a specific entity across the three datasets, with the order of entities randomized to prevent bias from consecutive judgments.
The evaluation questions focused on several critical aspects:

\vspace{-2mm}
\begin{itemize}[leftmargin=*]
\itemsep-0.5mm
\item \textbf{Familiarity and comprehensiveness}: Workers indicated their familiarity with the subtypes and whether all expected types were included.
\item \textbf{Categorization quality}: Workers evaluated the clarity of subtype, part, and material distinctions, and whether the listed parts were essential.
\item \textbf{Level of detail}: Workers judged if the data's granularity was appropriate, avoiding excessive technicality or overgeneralization.
\item \textbf{Consistency across classifications}: Workers checked for inconsistencies in the level of detail within the listed subtypes and parts.
\end{itemize}

\vspace*{-2mm}
Table \ref{table:dataset-comparison} provides insights into the strengths and limitations of the three datasets, in various response categories. Few-shot data demonstrates a higher mean score (88.39) compared to human-annotated data (85.08) and exhibits a balanced level of detail, offering classifications that are clearer and more distinct. Unlike zero-shot data, which tends to generate overly specific or redundant details, the few-shot model avoids overgeneralization while capturing subtle elements that human annotators might overlook. Specifically, human-annotated data struggled with clarity and distinction in the parts category due to omissions, such as failing to list essential components (e.g., a ballot box without a lid or locking mechanism). This suggests that subjective interpretation and variability may introduce inconsistencies in human annotations, whereas the few-shot model is more thorough and consistent. On the other hand, zero-shot data excels in subtype coverage (82.74) compared to human-annotated data (21.05), yet its overly specific classifications and redundant part generation lower its scores in other areas, such as clarity and focus on essential components. This comparison underscores the potential of few-shot learning to strike a practical balance between precision and completeness, bridging gaps left by human annotations and zero-shot learning.

\section{Conclusion}
Our study explores the potential and challenges of using LLMs to extract structured knowledge about the subtypes, parts, and materials of physical objects. We found that few-shot and zero-shot methods offer unique advantages, balancing specificity, coverage, and clarity in ways that sometimes exceed human-annotated data. This work advances efforts toward AI systems that are not only accurate, but also explainable, adaptable, and transparent.

\section*{Limitations}
While our proposed work demonstrates promising results in extracting and organizing part and material knowledge from LLMs, it is not without limitations. These limitations span several areas, including the nature of knowledge representation, learning techniques, domain specificity, computational scalability, and risks of inconsistency.

First, a fundamental limitation arises from the nature of the knowledge being extracted. Different people conceptualize the structure and attributes of objects in varying ways, making it difficult to achieve a unified, objective representation. For example, one person might classify a box kite as consisting of \textit{spars}, \textit{sails}, and \textit{bridle}, while another might describe it as having \textit{a rigid framework}, \textit{cover on multiple sides}, \textit{bridle}, and \textit{tethering line}. Between these classifications, only bridle is common, and determining which classification is ``correct'' is more philosophical than factual. Similarly, an iced tea spoon could be seen as having distinct parts (e.g., a bowl and a handle) or merely as a singular, inseparable object. Likewise, whether a pamphlet should be divided into \textit{front} and \textit{back}, given that nearly every physical object has such aspects, depends entirely on subjective framing. This inherent subjectivity complicates the generation and organization of data, as well as the evaluation of its quality.

Second, the use of few-shot and zero-shot learning presents trade-offs between specificity and generality. A promising direction for future work lies in adopting hybrid approaches, potentially using multi-step prompts with in-context learning. This would help maintain the hierarchical consistency of generated knowledge and ensure each step accounts for previously generated data, reducing redundancy and improving coherence.

Third, redundancy is another challenge observed in the current approach. Ideally, if a superordinate type possesses certain properties, its subtypes should inherit those attributes by default. However, the current method stores information independently at each level of the hierarchy, with no awareness of parent or child relationships unless explicitly specified. Without further refinements, this can lead to duplicated information across hierarchical levels.

Next, the method's focus on external physical attributes---specifically, parts and materials---limits its generalizability to other types of knowledge. The experiments primarily concentrate on these concrete attributes, leaving open the question of how well the approach would perform with more abstract or conceptual knowledge. Extending the method to other domains or knowledge categories will likely require additional adaptations in methodology.

Another limitation is the dependence on domain-specific expertise for crafting effective prompts. Writing precise prompts requires subject-matter knowledge, which may pose a barrier for users without expertise in the target domain. This limits the scalability and accessibility of the approach for non-expert users or broader applications.

Additionally, scalability issues might appear when working with more complex problems. For example, in our experiments, identifying parts and materials required up to eight multi-step zero-shot prompts. Addressing more complex or abstract questions would likely require even more prompts, introducing computational overhead and potentially degrading performance. Furthermore, such multi-step approaches can be resource-intensive, requiring significant GPU capacity, which may limit accessibility for smaller research teams or developers with limited computing resources. Although our experiments used GPT-4, it would have been beneficial to test the approach on smaller models to better understand its feasibility in low-resource settings.

Finally, relying on LLMs involves risks of hallucination and inconsistency. Despite efforts to ensure reliability, LLMs can generate contradictory or incorrect information. In our experiments, rephrasing the same question was sometimes necessary to maintain consistency. For example, the model identified an `open-end adjustable spanner' as a type of adjustable spanner. However, when asked a follow-up verification question, it corrected itself, stating that open-end spanners are not adjustable by definition, and that the correct term should be ``adjustable spanner'' or ``adjustable wrench.'' Such inconsistencies require additional validation efforts, adding to the overall complexity of using LLMs effectively.

\section*{Ethics Statement}
This research acknowledges the potential risks associated with the misuse of certain information. While the study includes the classification of objects such as weapons, caution is advised when accessing this data. Some details, such as the parts and materials of dangerous devices, could be exploited by individuals with harmful intent. However, in cases where the risk of misuse was deemed too high, the LLM deliberately withheld specific information. For example, the LLM listed `Suicide vest' as a subtype of `Explosive device,' but it refused to generate parts or material details to mitigate potential danger.\newline\newline
All code and data used and generated in this project are publicly  and freely accessible to promote transparency and reproducibility while encouraging responsible use.

\bibliography{anthology, custom}

\onecolumn
\appendix

\section{Constructing common physical objects}
\label{sec:common-objects-long}
Since we are interested in acquiring commonsense knowledge about the world, and the material composition of physical objects is an important aspect of such knowledge, our initial aim is to collect a list of physical objects that many people perceive or interact with. We are focusing on artifacts since their part structure, and especially their material composition, is generally clearer than that of natural objects. Consider for example humans, bushes, mountains, or stars; also, we plan to extend this work to provide usage (telic) information about objects, an aspect that applies much more clearly and consistently to artifacts than to natural objects.

For this, we first obtain all Wikidata entities from its JSON dump (downloaded on May 26th, 2023). We select entities that are subclasses of \textit{artificial physical objects, artificial physical structures}, or \textit{artificial entities}. The selected list still contains abstract entities such as `process' and `genre' due to multiple inheritances in Wikidata. To address this, we apply two layers of filtering:
\begin{enumerate}
\vspace{-0.3cm}
\item Keyword-based filtering: If the entity's Wikipedia title contains specific keywords, it is excluded. Examples of such keywords include law, protein, pattern, physics, topology, and unit.
\vspace{-0.3cm}
\item Subclass-based filtering: If an entity belongs to specific subclasses, it and all its descendant categories are excluded. Examples of these subclasses are concept, detection, genre, formal system, organism, and sound.
\vspace{-0.2cm}
\end{enumerate}

The list of entities is further filtered to ensure that it contains only commonly known nouns that occur as dictionary entries. As a dictionary source, we use the library module for WordNet, a lexical database, in NLTK (https://www.nltk.org). We filter out entries based on the availability of the entries and their synonyms in WordNet; we filter out entries if neither the entry noun itself nor any of its synonyms exist in WordNet, and if the entry does not have any link from Wikidata to its associated WordNet synset. Wikidata entries without a link to their corresponding Wikipedia articles are also excluded. Lastly, to refine the entries to include only physical, commonly encountered, and readily perceived objects, we ask GPT-4 a sequence of four filtering prompts. First, we use a prompt to identify readily perceived entities that a sixth-grader would know, then filter for physical objects. Next, we narrow the list to count nouns, and finally, ensure the remaining items are individual, standalone objects. The prompts in the filtering process can be found below.

\subsection{Prompt to identify readily perceived entities}
How likely are the following 50 things to be commonly recognized by a typical sixth-grader? Add ` - [likely / probably likely / probably unlikely / unlikely] to be recognized by sixth-graders' after the nouns in the list. Please do not alter the names within parentheses.
\subsection{Prompt to identify physical entities}
Could you classify the following 50 nouns based on whether they primarily refer to standalone physical objects, standalone built structures, substances, or neither? Add ` - is a [physical object / built structure / substance / neither]' after the nouns in the list. Please do not alter the names within parentheses.
\newline\newline
Here are the criteria for each category:\\
- Physical objects: Tangible items that can exist independently, or items that might be part of a larger entity but can be replaced.\\
- Built structures: Man-made constructions that serve as physical places or infrastructure.\\
- Substances: Any substance with uniform characteristics or any matter that can be best characterized by their chemical composition.\\
- Neither: None of the above.
\subsection{Prompt to identify count nouns}
Could you classify the following 50 nouns based on whether they are in general used as a mass noun or a count noun? Add ` - mass noun' or ` - count noun' accordingly after the nouns in the list.
\subsection{Prompt to identify individual standalone entities}
Could you classify the following 50 nouns based on whether they are typically described as an entity on their own, or composed of multiple standalone entities? Add ` - a single entity', ` - a group of components but commonly referred to as a single item', or ` - a group of multiple standalone items' accordingly after the nouns in the list.

\section{Few-shot in-context learning prompts}
\label{sec:fewshot-prompt}
\subsection{Prompt for subtypes and parts}
Please list common categories and their sub-categories, and their constituent parts of the given entity. Each type must be distinguished solely by the unique presence of their essential parts or components. Only list essential parts, not in their variations in shape, size, material, or function. Please do not count chemical substances such as electrolyte as essential parts.
\newline\newline
Alternatively, you may state ``No distinct subtypes based on the constituent parts'' instead of listing subtypes if there are no variations in the essential, unique parts that distinguish the subtypes. Then, indicate ``Physical parts'' underneath.
\newline\newline
Please do not state any descriptive terms or clarifications within parentheses. Only indicate ``(optional)'' where applicable. You may use "internal mechanism" as a part for any components not visible externally but essential for function.
\newline\newline
\#\#\newline
Entity 1: Barn\newline
Subtypes 1:\newline
1. English barn: walls, roof, floor, frame, three bays\newline
2. Livestock barn: walls, roof, floor, frame, tack room, feed room (optional), drive bay, silo, stalls\newline
3. Dairy barn: walls, roof, floor, frame, tack room, feed room, drive bay, silo, stalls, milk house, grain bin, indoor corral (optional)\newline
4. Crop storage barn: walls, roof, frame, drive bay\newline
5. Crib barn: walls, roof, cribs, roof shingles\newline
6. Bank barn\newline
6.a) New England barn: walls, roof, roof shingles, floor, tack room (optional), frame\newline
6.b) Pennsylvania barn: walls, roof, roof shingles, floor, forbear, frame, gables (optional)\newline
\#\#\newline
Entity 2: Saucer\newline
Subtypes 2: No distinct subtypes based on the constituent parts.\newline
Physical parts: No distinct parts\newline
\#\#\newline
Entity 3: Paintbrush\newline
Subtypes 3: No distinct subtypes based on the constituent parts.\newline
Physical parts: handle, bristle, ferrule\newline
\#\#\newline
Entity 4: Frying pan\newline
Subtypes 4:\newline
1. Stovetop frying pan: body, handle\newline
2. Electric frying pan: body, handle, legs, lid (optional), lid knob (optional), power cord, thermostat\newline
\#\#\newline
Entity 5: Glove (ice hockey)\newline
Subtypes 5:\newline
1. Skater's gloves: palm, back, fingers, padding\newline
2. Blocker: palm, back, fingers, padding, forearm pad\newline
3. Trapper: palm, back, fingers, padding, cuff, pocket, inner glove\newline
\#\#\newline
Entity 6: \texttt{[noun]}\newline
Subtypes 6:

\subsection{Prompt for materials}
Please list the materials that the listed parts of the given entity are typically made of. Exclude any materials used for joining, stitching or dying.
\newline\newline
Allow any necessary repetition in materials across different parts. Avoid using ``sometimes'', ``such as'', and parentheses in your response. Connect the materials with one of the following conjunctions:\newline
- ``and'': all listed materials are typically used together\newline
- ``or'': each of the materials from the list is used exclusively\newline
- ``and/or'': some of the listed materials are typically used in combination
\newline\newline
\#\#\newline
Entity 1: Peripheral webcam (Webcam)\newline
Parts: case, camera lens, image sensor, mount, interface\newline
Materials:\newline
1. case: plastic\newline
2. camera lens: plastic or glass\newline
3. image sensor: electronics\newline
4. mount: metal\newline
5. interface: electronics, metal, and plastics\newline
\#\#\newline
Entity 2: Paper cup\newline
Parts: cup, cardboard lining, lid\newline
Materials:\newline
1. cup: paper\newline
2. cardboard lining: plastic or wax\newline
3. lid: plastic\newline
\#\#\newline
Entity 3: Facial tissue\newline
Parts: -\newline
Materials: absorbent paper\newline
\#\#\newline
Entity 4: Paintbrush\newline
Parts: bristle, ferrule, handle\newline
Materials:\newline
1. bristle: animal hair, nylon, and/or polyester\newline
2. ferrule: metal\newline
3. handle: wood or plastic\newline
\#\#\newline
Entity 5: Skater's gloves (ice hockey)\newline
Parts: palm, back, fingers, padding\newline
Materials:\newline
1. palm: leather\newline
2. back: leather and/or kevlar\newline
3. fingers: leather and/or kevlar\newline
4. padding: foam\newline
\#\#\newline
Entity 6: \texttt{[noun]}\newline
Parts: \texttt{[parts]}\newline
Materials:

\section{Zero-shot in-context learning prompts}
\label{sec:zeroshot-prompt}
\subsection{Prompt to assess whether an entity has subtypes}
Are there any essential, non-optional parts\\
1) that are present in one type of \texttt{[noun]} but absent in another and\\
2) that would be recognized by most people?\\
Simply say ``yes'' or ``no''.

\subsection{Prompt to generate a list of subtypes}
In numbered points, please simply list physically distinct types of \texttt{noun}, where each type is distinguished by unique, externally visible, essential parts.\newline\newline
Exclude any categories that share the same essential external components and functions. The listed categories should reflect differences in their primary operation rather than just external design variations or connections.\newline\newline
Also, avoid from your list any categories that merely represent design variations, subtypes, or alternate names for the same tool. Format each entry as a complete noun without using `traditional', `and', `or', nouns indicating materials, or any prepositional phrases such as `with' in the names.

\subsection{Prompt to identify common subtypes}
How likely would the following types of \texttt{noun} be recognized by most people? Add `` - [likely / probably likely / probably unlikely / unlikely] recognized by most people'' after the nouns in the list. Please do not alter the names within parentheses.

\subsection{Prompt to assess whether an object has parts}
How many parts does \texttt{noun} have? Specifically, how many clearly distinct parts that are attached to it or inseparable from it? Please simply say the number of parts.

\subsection{Prompt to assess whether an object has uniform materials across different parts}
Are distinct parts of \texttt{noun} made of the same materials? Say ``yes'' or ``no''.

\subsection{Prompt to identify materials of an object}
In one line, please list solely the types of materials that \texttt{noun} are typically made of. Avoid using ``sometimes'', and connect the materials with a conjunction, e.g., `glass, plastic, and/or metal'. Exclude any materials used for joining, stitching or dying.\newline\newline
Here are the conjunctions you can use:\\
- ``and'': all listed materials are typically used together\\
- ``or'': each of the materials from the list is used exclusively\\
- ``and/or'': some of the listed materials are typically used in combination.'

\subsection{Prompt to identify parts and materials of an object}
1) Starting your paragraph with ``<Parts>\textbackslash n'', in numbered points, please list clearly distinct, essential parts of \texttt{noun} with succinct descriptions followed by ``:''. For each part, insert a new line that starts with ``- Optional:''. Answer with ``yes'' or ``no''.\newline\newline
2) Starting your paragraph with ``<Materials>: '', in new bullet points, please list solely the materials that a typical \texttt{noun} is entirely made of. Avoid using ``sometimes'', and connect the materials with a conjunction, e.g., `<Materials>: glass, plastic, and/or metal'. Exclude any materials used for joining, stitching or dying. Here are the conjunctions you can use.
- ``and'': all listed materials are typically used together\\
- ``or'': each of the materials from the list is used exclusively\\
- ``and/or'': some of the listed materials are typically used in combination.\newline\newline
Keep your answers very simple, in terms a second-grader would understand.

\subsection{Prompt to identify an object's parts and their materials}
In numbered points, please list the clearly distinct, essential parts of \texttt{noun} that are attached to it or inseparable from it, with succinct descriptions following ``:''. Things that have multiple independent uses, such as `battery', don't count as a part. You may use ``internal mechanism'' as a part for anything that is not visible from the outside.\newline\newline
For each part, insert a new line that starts with ``- Optional:''. Answer with ``yes'' or ``no''.\newline\newline
Then again, for each part, insert a new line that starts with ``- Materials:'' and mention the materials the part is typically made of. List the materials, avoiding using ``sometimes'', and connect the materials with a conjunction, e.g., `- Materials: glass, plastic, and/or metal'. Here are the conjunctions you can use.\\
- ``and'': all listed materials are typically used together\\
- ``or'': each of the materials from the list is used exclusively\\
- ``and/or'': some of the listed materials are typically used in combination.\newline\newline
Keep your answers very simple, in terms a second-grader would understand.

\section{Overview of the acquired subtypes, parts, and materials}
\label{sec:overview-distribution}
The acquired datasets are hierarchically organized into entities (Wikipedia entries), subtypes, and subsubtypes, with parts and materials. At the most granular level, individual items at the lowest level in the hierarchy are characterized by their constituent parts and materials. If certain items lack subtypes or subsubtypes, the parts and materials associated with the higher-level entity are listed instead.

\begin{itemize}
    \item \textbf{Total items}: The occurrences of items that contain associated parts and/or materials, across all entities.
    \begin{itemize}
        \item Few-shot: 6,275 items
        \item Zero-shot: 27,285 items
    \end{itemize}
\end{itemize}

\subsection{Distribution of subtypes and subsubtypes}
\begin{itemize}
    \item \textbf{Entities without subtypes}: the number of entities are classified without distinct subtypes.
    \begin{itemize}
        \item Few-shot: 1,167 entities
        \item Zero-shot: 677 entities
    \end{itemize}
    
    \item \textbf{Few-shot subtypes}
    \begin{itemize}
        \item 4,851 unique subtyeps all entities
        \item 5,056 unique subtypes per entity
        \item 84 unique subsubtypes
    \end{itemize}

    \item \textbf{Zero-shot subtypes}
    \begin{itemize}
        \item 8,486 unique subtyeps across all entities
        \item 8,987 occurrences of subtypes per entity
        \item 21,329 unique subsubtypes across all entities
        \item 22,489 unique subsubtypes per entity
        \item 22,780 unique subsubtypes for every entity and each of its subtypes 
    \end{itemize}
\end{itemize}

\noindent The difference in the number of unique occurrences of subtypes or subsubtypes is due to overlapping entries. Below is an example of such overlaps:
\begin{itemize}
    \item Kitchen utensil > Ladle > Soup ladle
    \vspace{-2mm}
    \item Spoon > Ladle > Soup ladle
    \vspace{-2mm}
    \item Spoon > Serving spoon > Soup ladle
\end{itemize}

\subsection{Distribution of parts and materials}
\begin{itemize}
    \item \textbf{Few-shot items without parts}
        \begin{itemize}
            \item 190 items (out of 1,167) lack parts
            \item 104 subtypes (out of 5,024) lack parts
            \item 3 subsubtypes (out of 84) lack parts
        \end{itemize}
    
    \item \textbf{Zero-shot items without parts}
        \begin{itemize}
            \item 92 items (out of 677) lack parts
            \item 207 subtypes (out of 3,828) lack parts
            \item 360 subsubtypes (out of 22,780) lack parts
        \end{itemize}

    \item \textbf{Average number of parts}
        \begin{itemize}
            \item Few-shot: 4.36 parts per item
            \item Zero-shot: 8.14 parts per item
        \end{itemize}

    \item \textbf{Average number of materials}
        \begin{itemize}
            \item Few-shot: 2.04 materials per item
            \item Zero-shot: 2.37 materials per item
        \end{itemize}
\end{itemize}

\section{Sample questions on Amazon Mechanical Turk}
\label{sec:mturk-samples}
\subsection{Precision}
\subsubsection*{Subtype precision}
\fbox{\parbox{0.80\textwidth}{%
\includegraphics[width=0.85\textwidth]{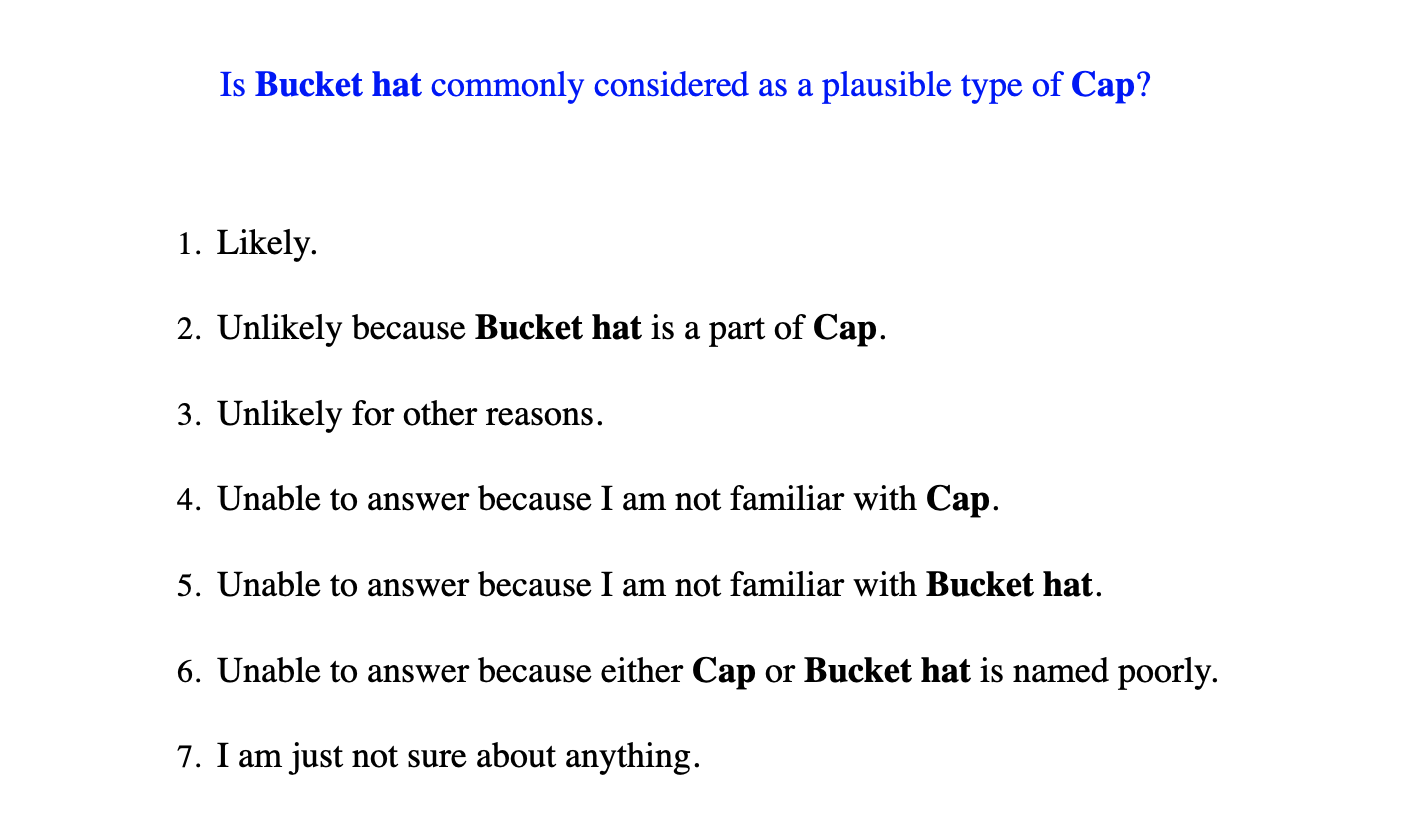}}}

\subsubsection*{Part precision}
\fbox{\parbox{0.80\textwidth}{%
\includegraphics[width=0.78\textwidth]{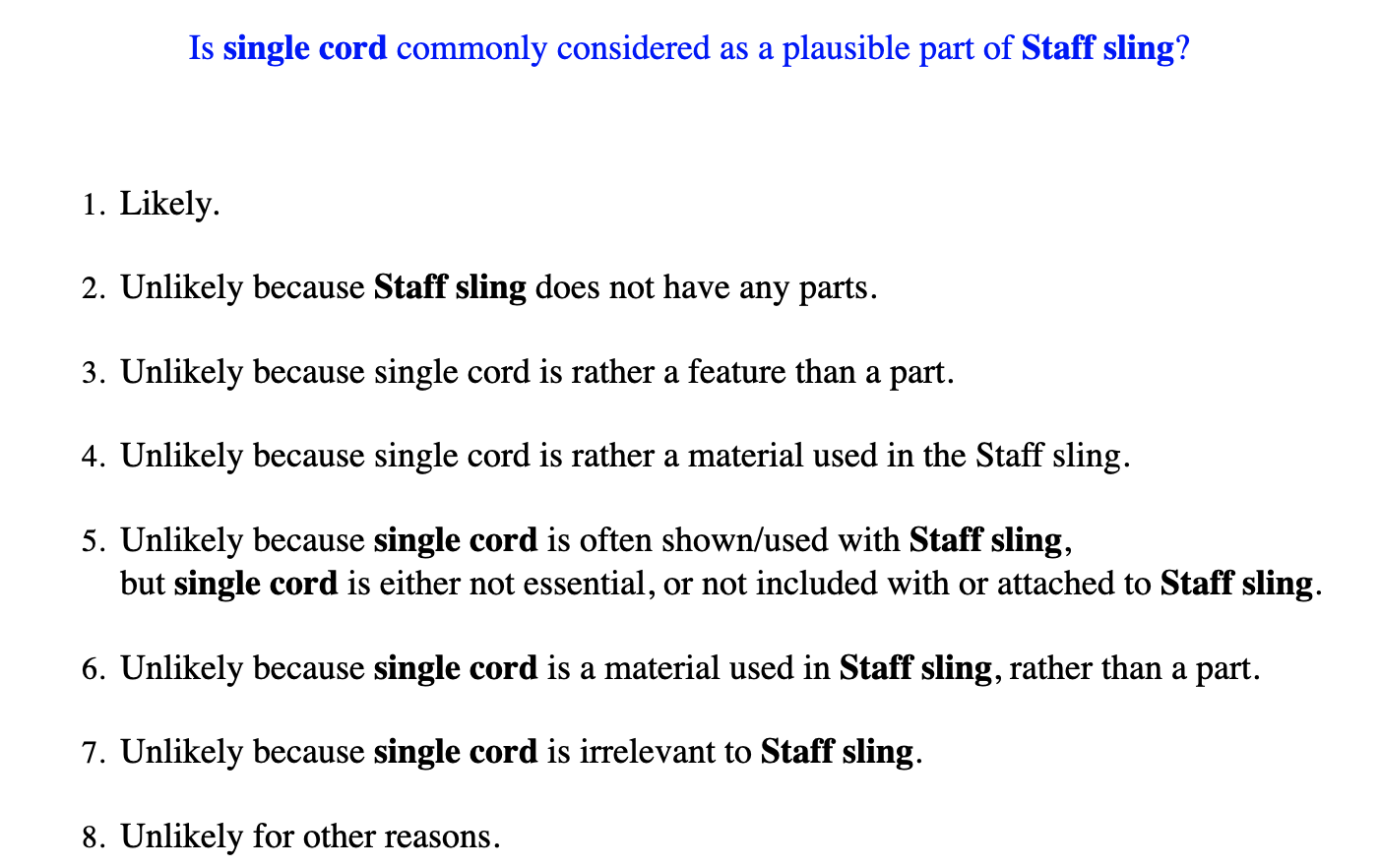}\\
\includegraphics[width=0.78\textwidth]{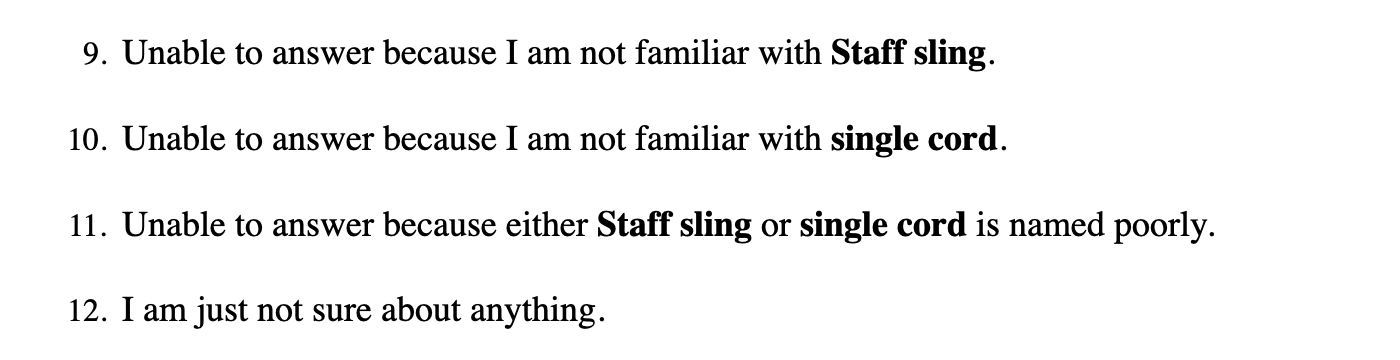}}}
\vspace{1cm}
\subsubsection*{Material precision}
\fbox{\parbox{0.95\textwidth}{%
\includegraphics[width=0.95\textwidth]{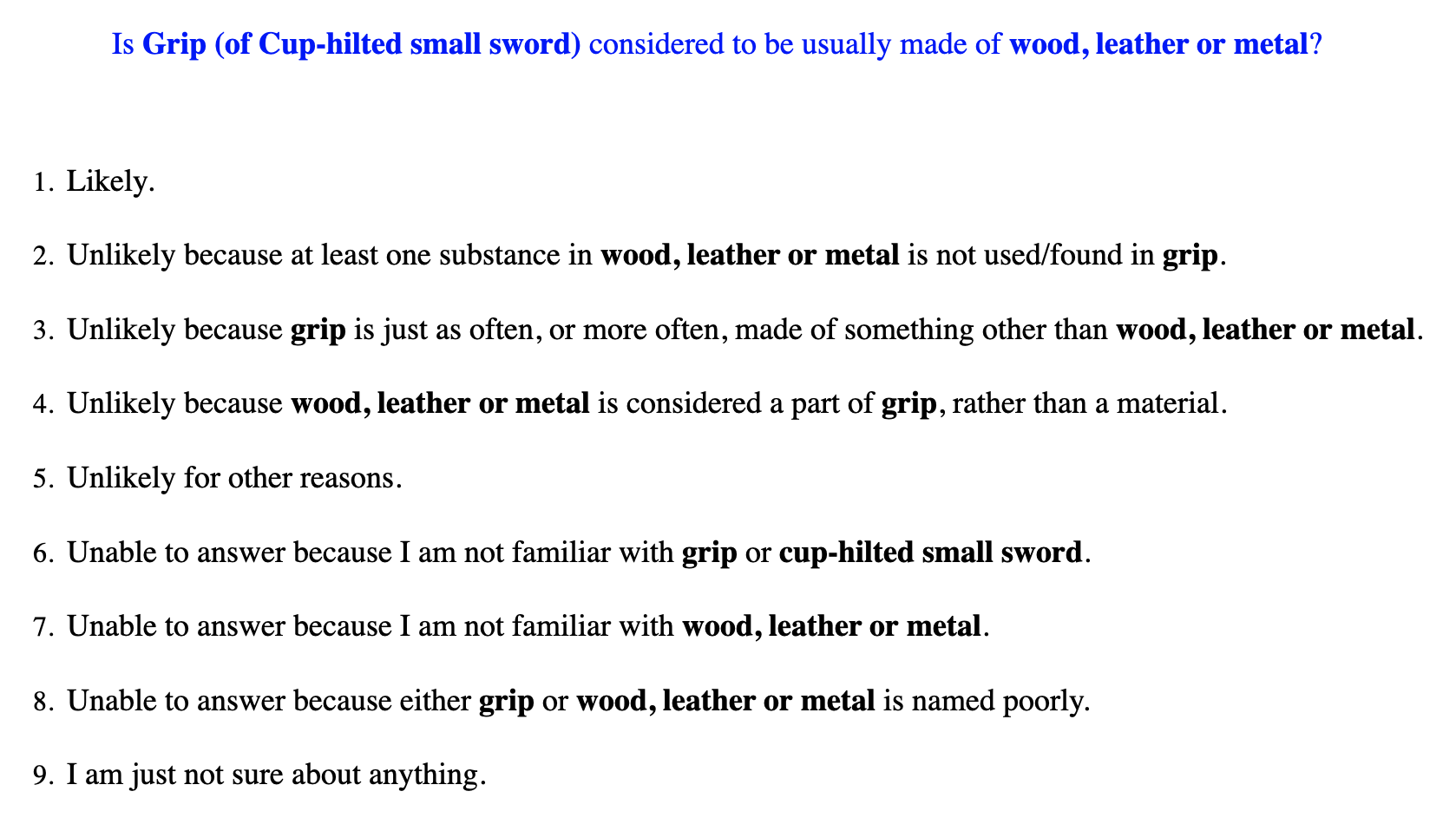}}}

\clearpage

\subsection{Recall}
\subsubsection*{Subtype recall}
\fbox{\parbox{0.85\textwidth}{%
\includegraphics[width=0.85\textwidth]{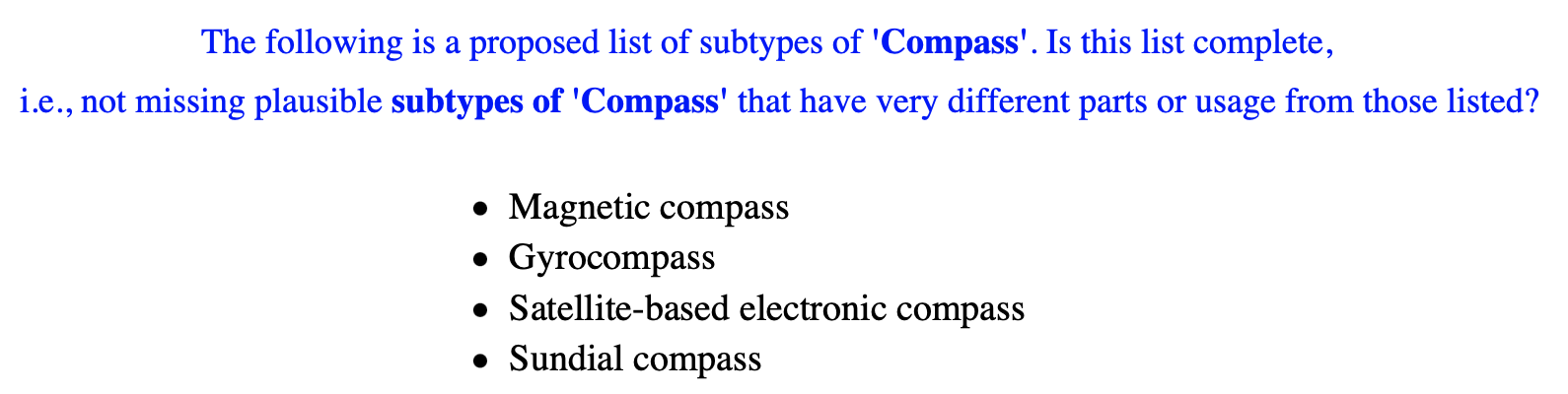}\\
\includegraphics[width=0.85\textwidth]{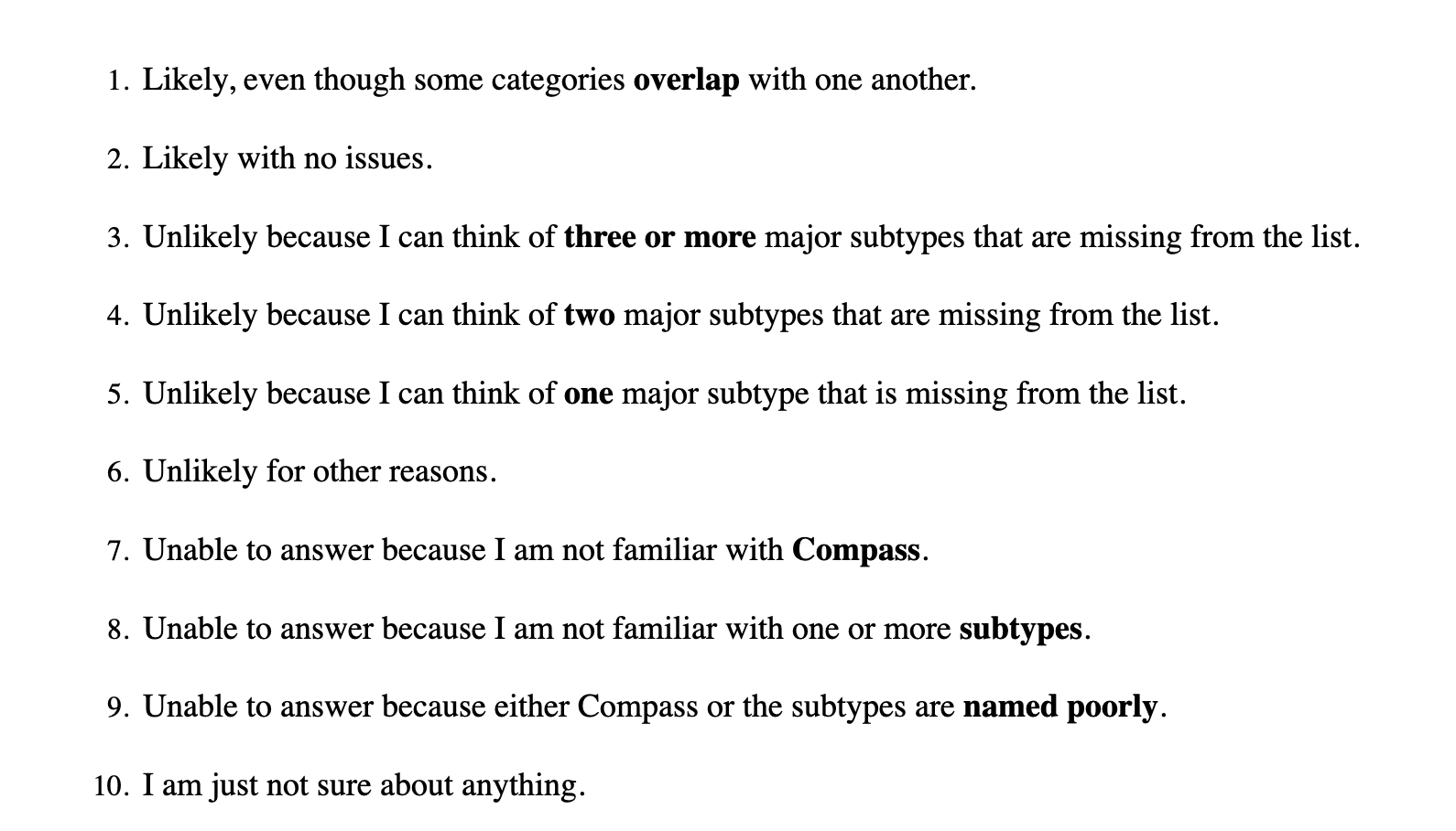}
}}

\subsubsection*{Part recall}
\fbox{\parbox{0.93\textwidth}{%
\includegraphics[width=0.93\textwidth]{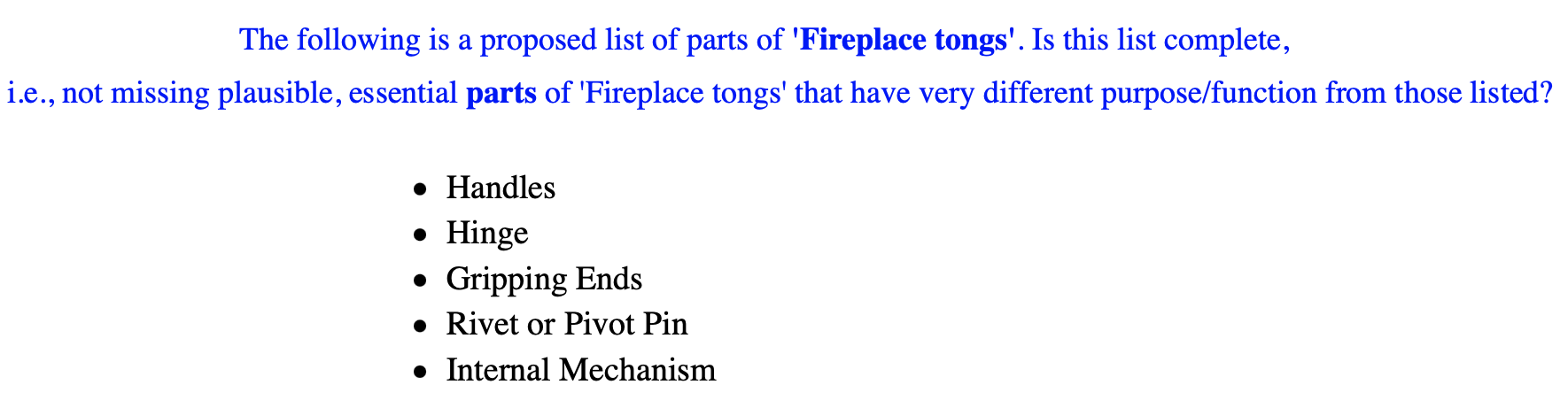}\\
\includegraphics[width=0.93\textwidth]{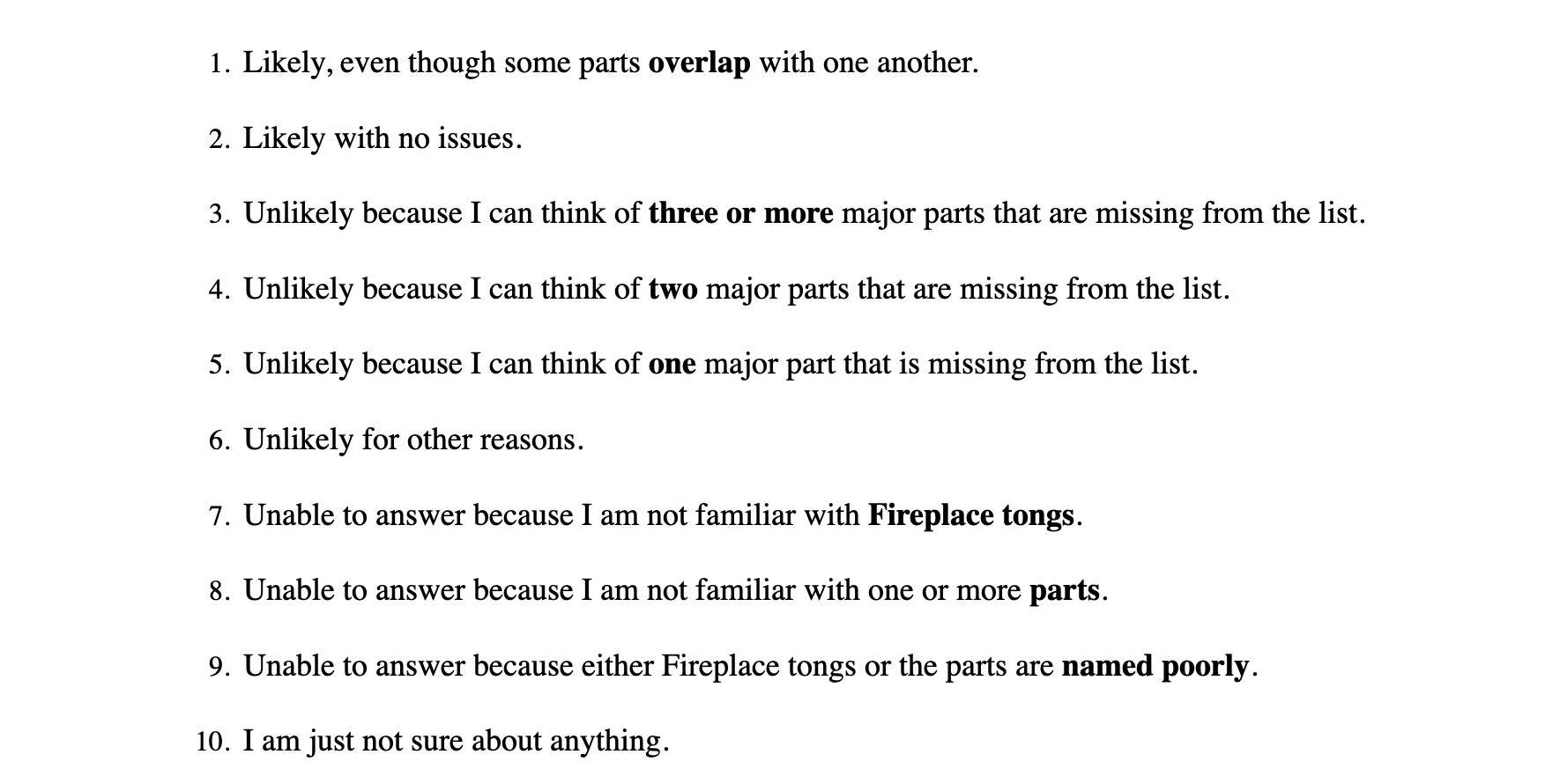}
}}

\clearpage 

\subsubsection*{Material recall}
\fbox{\parbox{0.98\textwidth}{%
\includegraphics[width=0.98\textwidth]{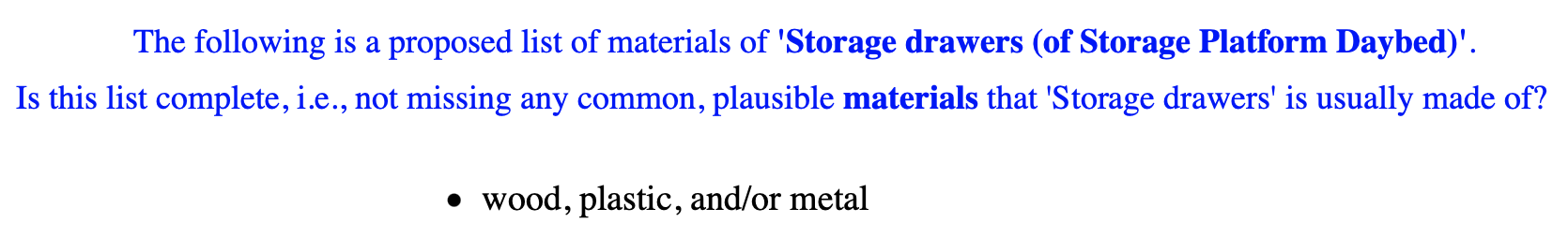}\\
\includegraphics[width=0.98\textwidth]{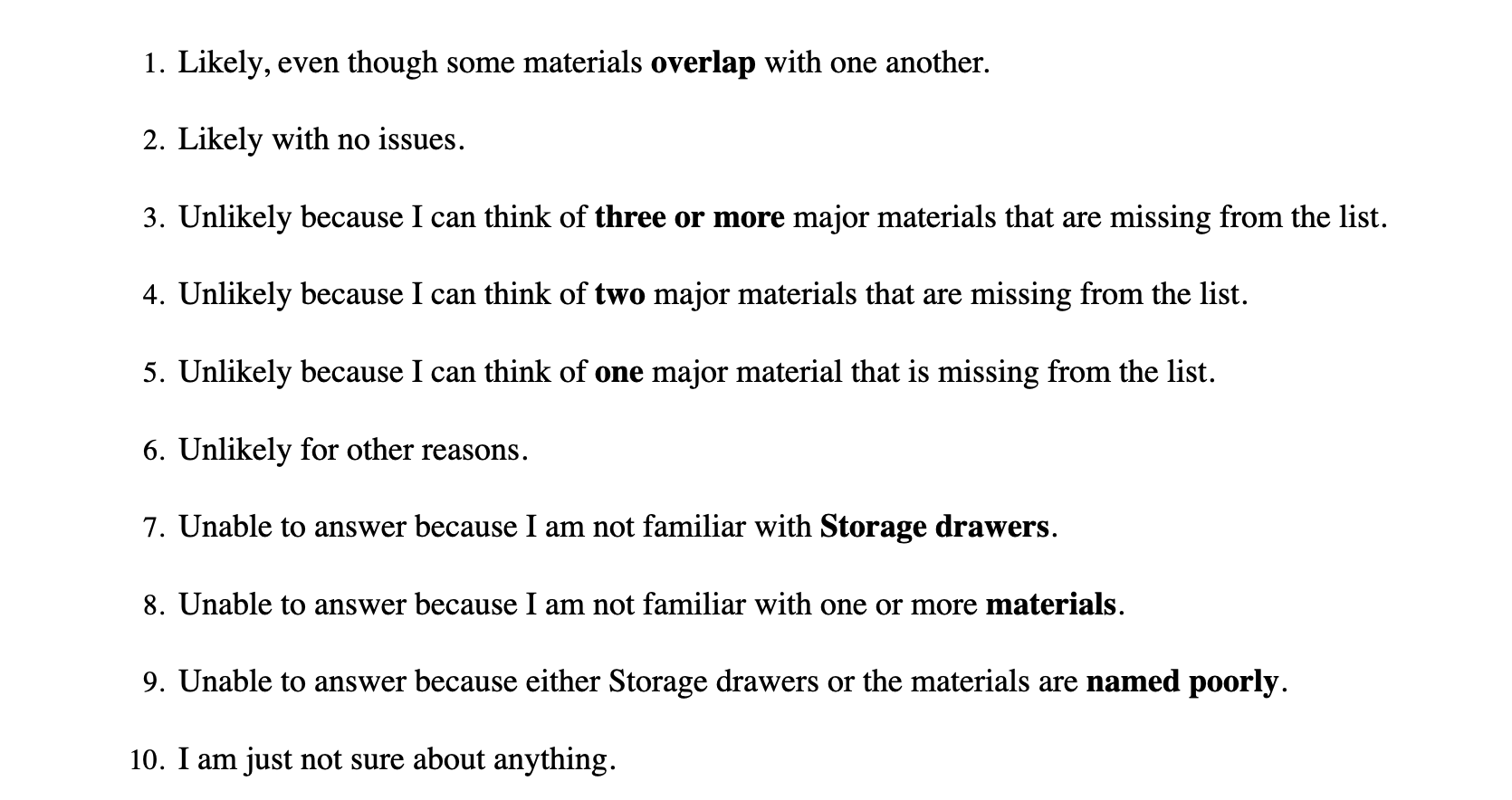}
}}
\vspace{1cm}
\subsection{Distinctive feature significance}
\fbox{\parbox{0.75\textwidth}{%
\includegraphics[width=0.75\textwidth]{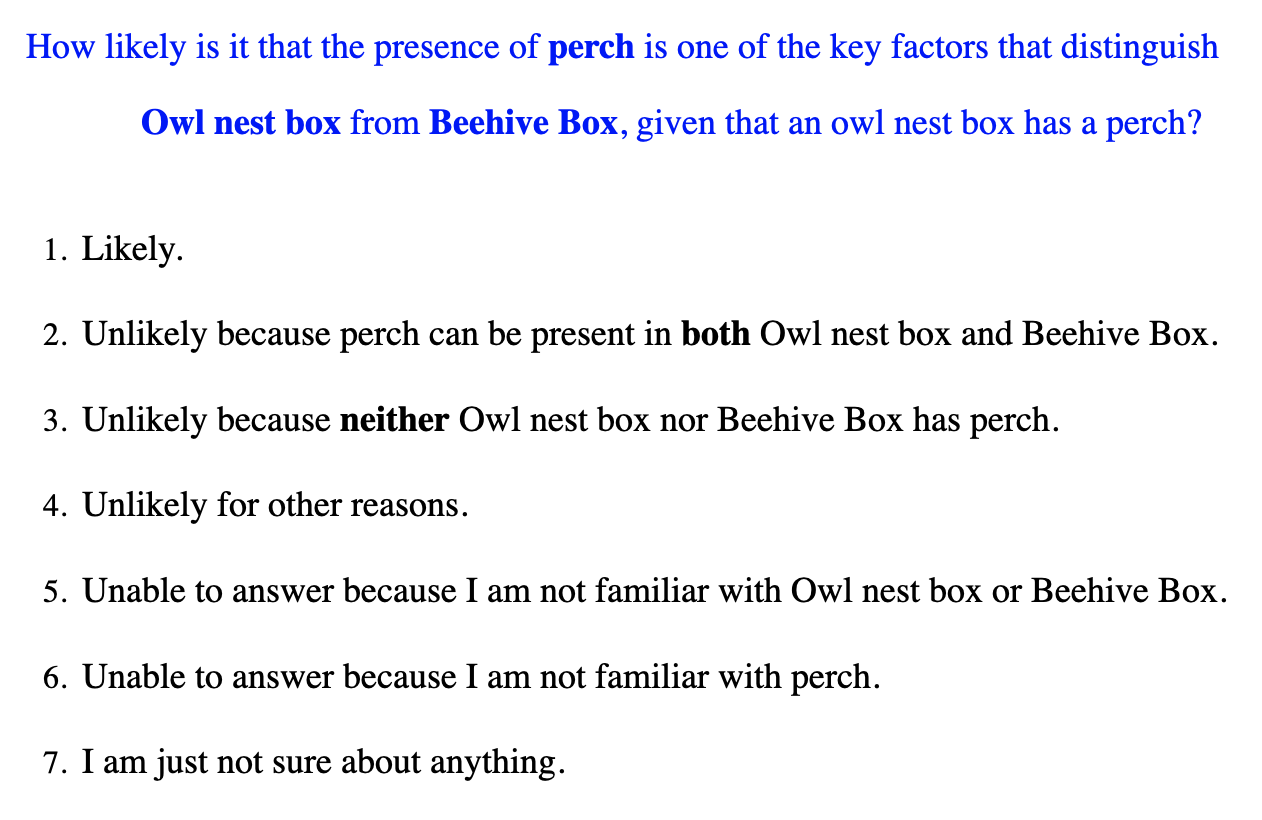}}}

\clearpage

\subsection{Dataset-comparison evaluation}
\fbox{\parbox{0.98\textwidth}{%
\includegraphics[width=0.98\textwidth]{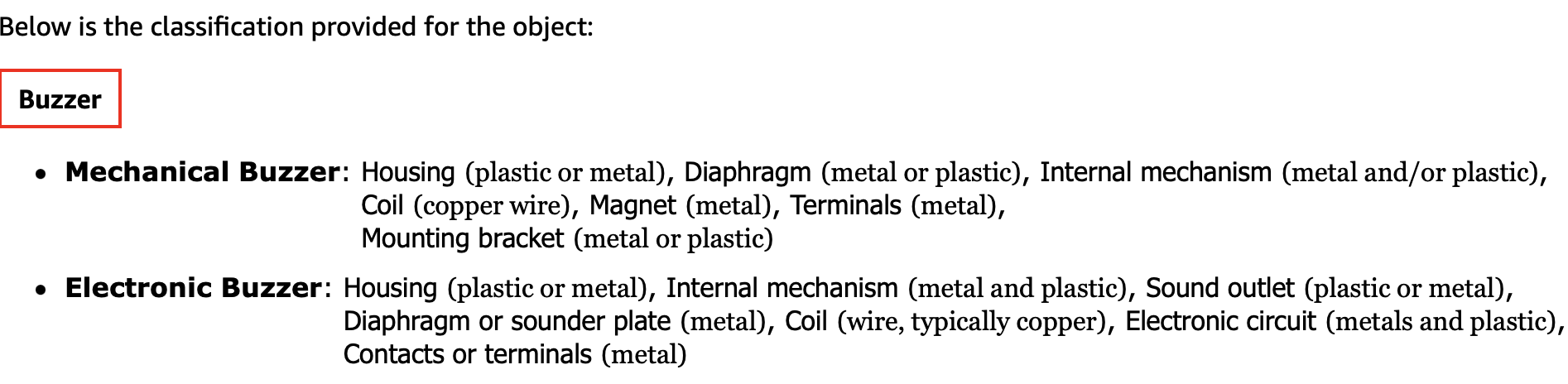}\\
\includegraphics[width=0.98\textwidth]{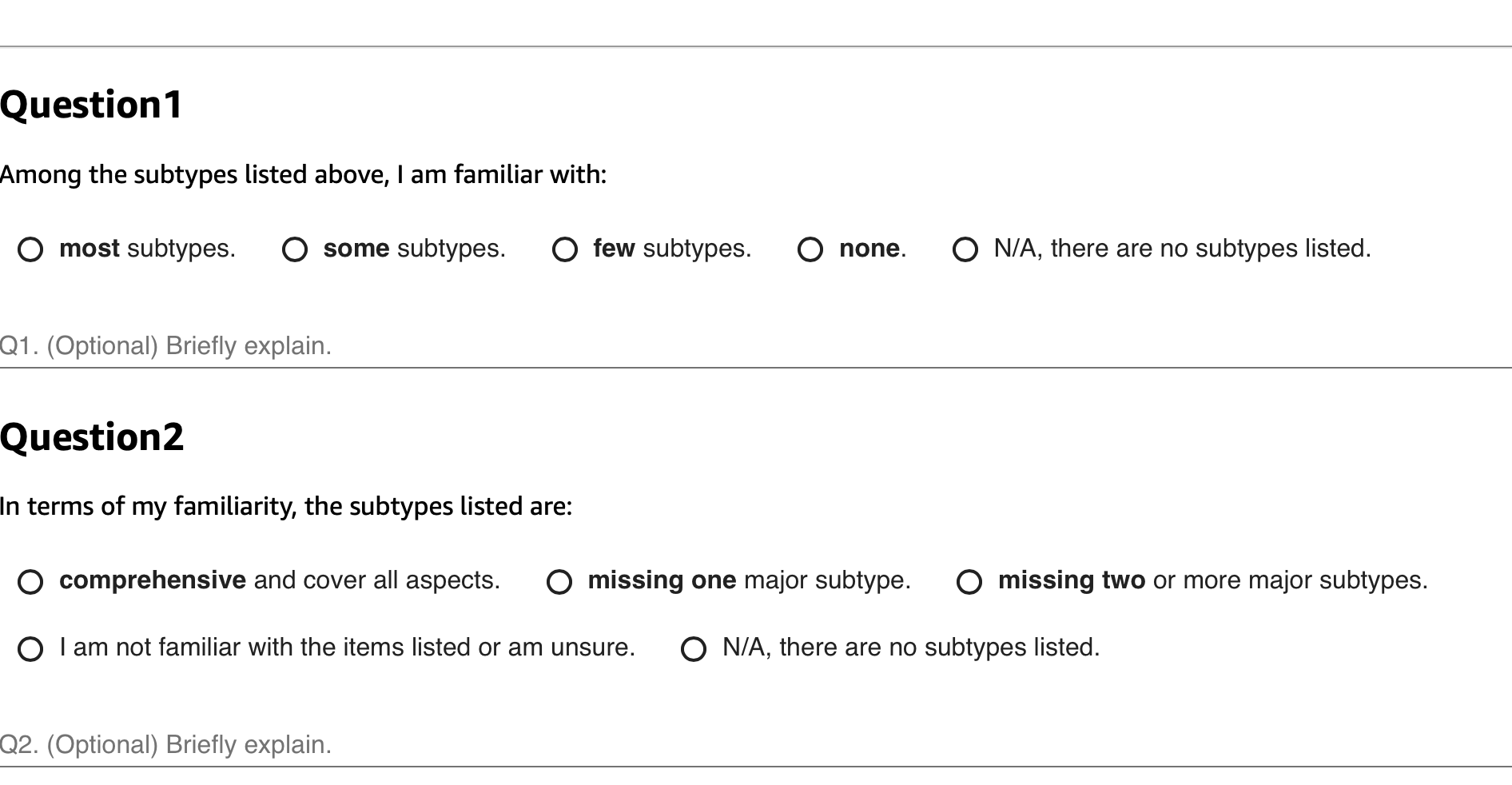}\\
\includegraphics[width=0.98\textwidth]{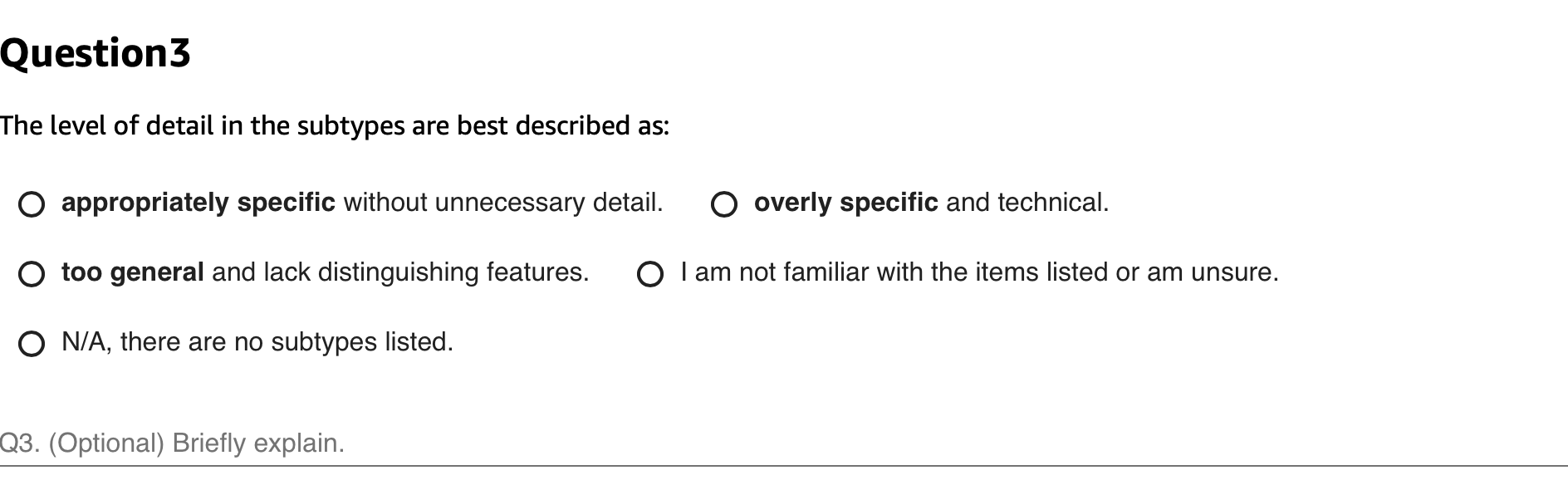}\\
\includegraphics[width=0.98\textwidth]{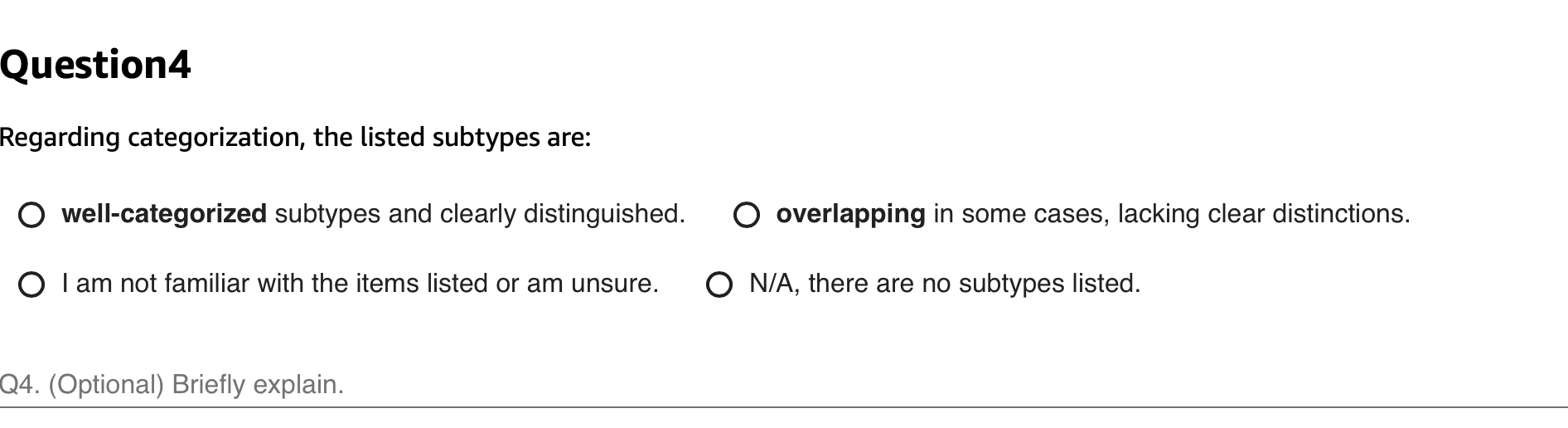}
}}

\clearpage

\fbox{\parbox{0.98\textwidth}{%
\includegraphics[width=0.98\textwidth]{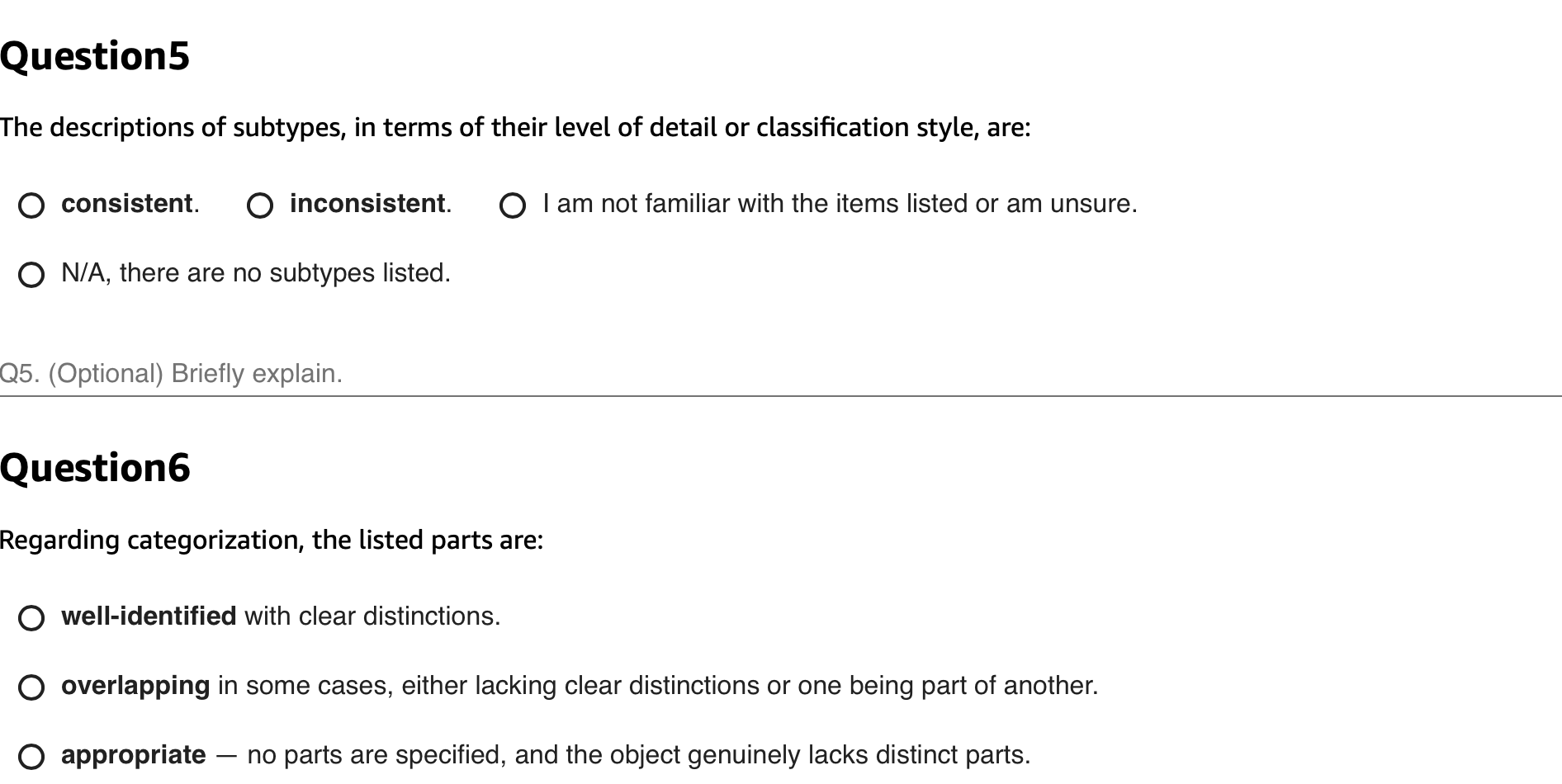}\\
\includegraphics[width=0.98\textwidth]{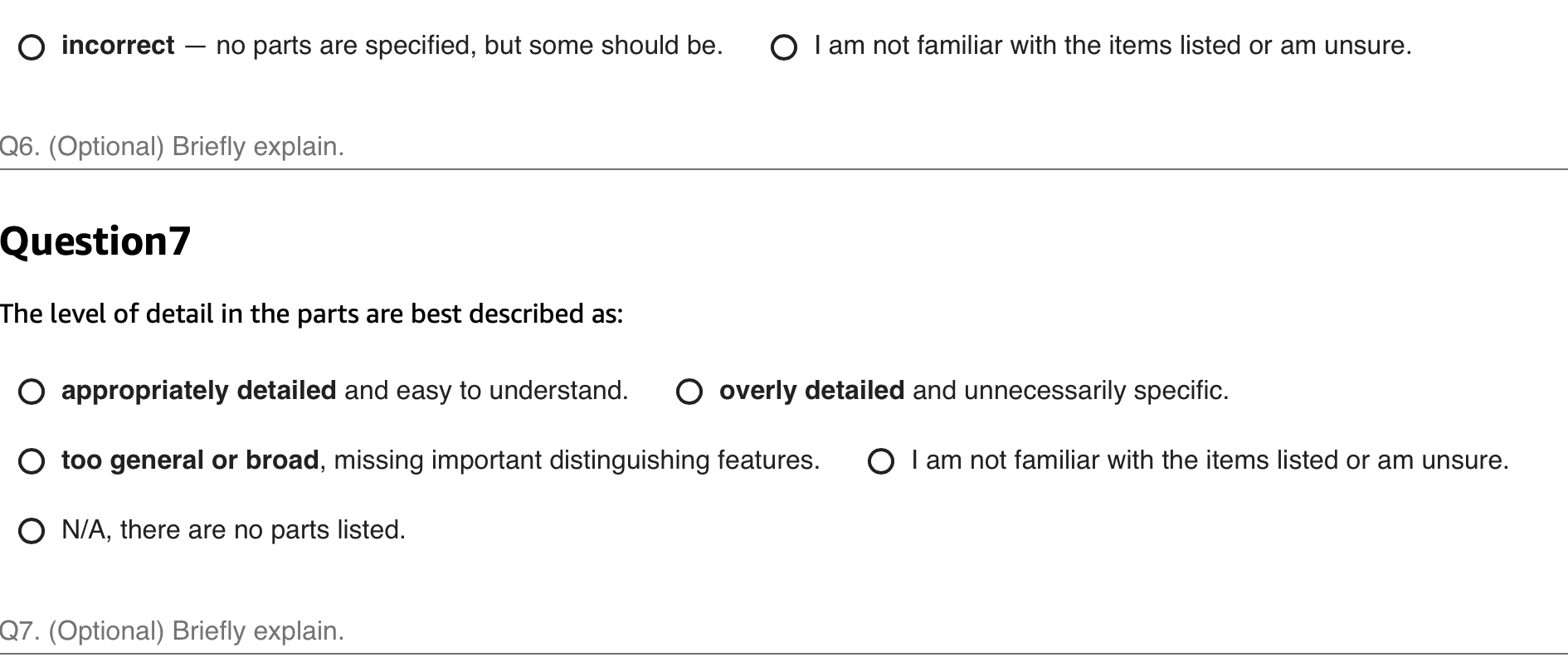}\\
\includegraphics[width=0.98\textwidth]{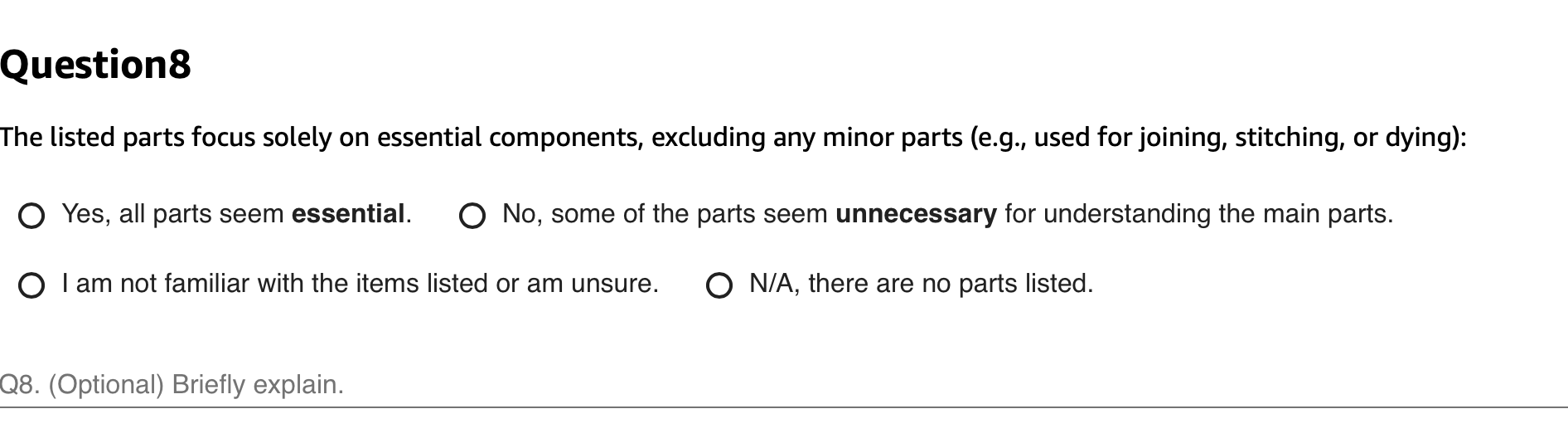}
}}

\clearpage

\fbox{\parbox{0.98\textwidth}{%
\includegraphics[width=0.98\textwidth]{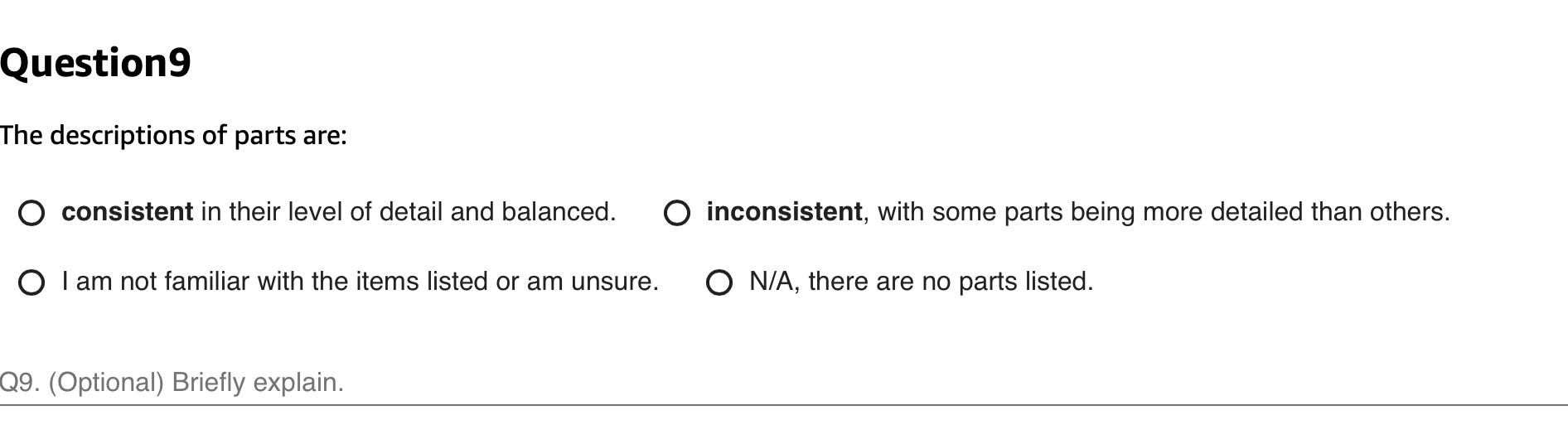}\\
\includegraphics[width=0.98\textwidth]{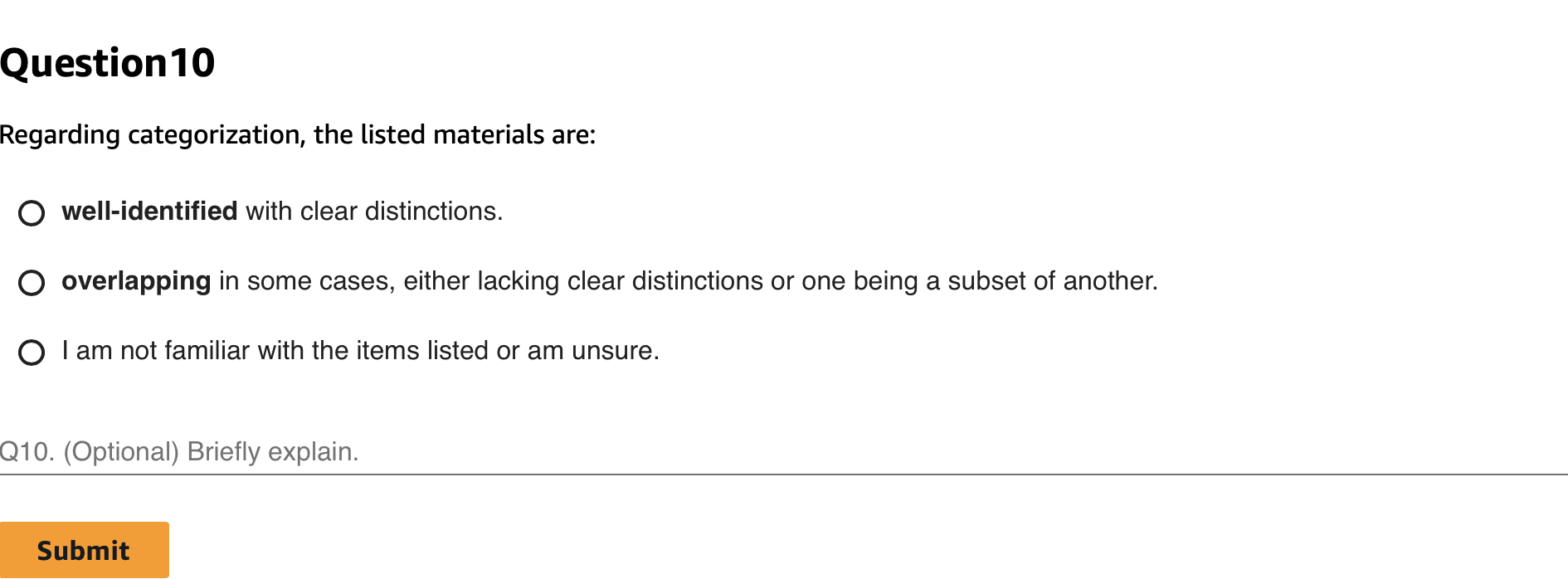}
}}

\clearpage

\section{Additional experiment results}
\label{sec:experiment-results}

\subsection{Breakdown of recall evaluation results based on external datasets}
\label{sec:external-recall-breakdown}
\begin{table*}[!thbp]
  \centering
  \begin{tabular}{@{\hskip 0.04in}c@{\hskip 0.12in}c@{\hskip 0.14in}c@{\hskip 0.15in}cccccc}
    \hline
    \textbf{Category} & \textbf{Our data} & \textbf{Reference data} & \textbf{Full-credit} & \textbf{Half-credit} & \textbf{Missing} &\textbf{Total} & \textbf{Recall (\%)}\\
    \hline
    &         & \hspace{0.25cm}ParRoT & 83 & 26 & 55 & 164 & 58.54\\
    &         & \hspace{0.25cm}CSLB & 63 & 13 & 30 & 106 & 65.57\\
    Part & Few-shot & \hspace{0.25cm}McRae & 18 & 1 & 5 & 24 & 77.08 \\
    &         & \hspace{0.25cm}WordNet & 37 & 6 & 10 & 53 & 75.47\\
    &         & \hspace{0.25cm}ConceptNet & 23 & 1 & 6 & 30 & 78.33\\
    \hline
    &          & \hspace{0.25cm}ParRoT & 151 & 17 & 6 & 174 & 91.67\\
    &          & \hspace{0.25cm}CSLB & 101 & 7 & 1 & 109 & 95.87 \\
   Part & Zero-shot & \hspace{0.25cm}McRae & 24 & 0 & 1 & 25 & 96.00\\
    &          & \hspace{0.25cm}WordNet & 44 & 5 & 4 & 53 & 87.74\\
    &          & \hspace{0.25cm}ConceptNet & 28 & 1 & 1 & 30 & 95.00\\
    \bottomrule
    \multirow{4}{*}{Material} & \multirow{4}{*}{Few-shot} & \hspace{0.25cm}CSLB & 59 & 1 & 7 & 67 & 88.81\\
    & & \hspace{0.25cm}McRae & 28 & 1 & 0 & 29 & 98.28\\
    & & \hspace{0.25cm}WordNet & 10 & 3 & 2 & 15 & 76.67\\
    & & \hspace{0.25cm}ConceptNet & 12 & 1 & 0 & 13 & 96.15\\
    \hline
    \multirow{4}{*}{Material} & \multirow{4}{*}{Zero-shot} & \hspace{0.25cm}CSLB & 68 & 0 & 2 & 70 & 97.14\\
    & & \hspace{0.25cm}McRae & 29 & 1 & 0 & 30 & 98.33\\
    & & \hspace{0.25cm}WordNet & 14 & 1 & 0 & 15 & 96.67\\
    & & \hspace{0.25cm}ConceptNet & 12 & 1 & 0 & 13 & 96.15\\
  \end{tabular}
  \caption{Summary of part and material availability and recall for external datasets. The table includes the number of parts and materials from other datasets that are categorized as present in our dataset with full credit (1), half credit (0.5), or missing (0). Recall is calculated as the ratio of available parts to the total number of parts and materials from other datasets.}
  \label{table:external-recall-breakdown}
\end{table*}

\subsection{\textit{Unlikely} responses in material precision}
\label{sec:internal-precision-material-breakdown}

\begin{table}[hbp]
  \centering
  \begin{tabular}{llc}
    \hline
    \textbf{Unlikely answer detail} & \textbf{Few} & \textbf{Zero}\\
    \hline
    Substance not found in item & 30.34 & 45.28\\
    Often made of something else & 16.85 & 29.25\\
    Material is considered a part & 48.31 & 18.87\\
    Other reasons & 4.49 & 6.60\\
  \end{tabular}
  \caption{Distribution of reasons for \textit{Unlikely} responses in precision evaluation of `material' items, showing the proportion of responses for each reason.}
  \label{table:internal-precision-material-breakdown}
\end{table}

\noindent Table \ref{table:internal-precision-material-breakdown}
outlines the reasoning behind the \textit{Unlikely} responses in the precision evaluation of materials. For few-shot data, the primary reason is that the listed items are considered ``parts'', rather than materials (48.31\%). This indicates that the LLM often confused components of an object with the actual materials. For example, materials such as ``plastic and/or LED bulbs'' are listed for the lights of a French refrigerated produce trailer, ``plastic, metal, and/or electronic components'' for the timer of a built-in electric grill, and ``plastic, metal, and speaker components'' for the Bluetooth speaker of an electric unicycle. In these cases, components like LED bulbs and electronic parts are not purely materials but rather a combination of materials and functional elements. This likely occurs because the part in question was too complex or its exact material composition is difficult to specify.

In contrast, for the zero-shot data, the most frequent reason is that at least one of the substances listed as materials is not found or used in the item (45.28\%). For instance, one worker responded a latch/clip part of a portable charcoal grill was unlikely to be made of metal or plastic. This might stem from the assumption that plastic would not be used in parts exposed to high heat, a logical inference given the grill's primary function. However, it's important to note that not every component of a grill is near the heat source, and some clips or latches may indeed be made of plastic. Thus, while the judgment that plastic parts are unlikely was reasonable, it does not entirely rule out the use of plastic in such components.

That said, as shown in Table \ref{table:internal-precision} and Table \ref{table:internal-precision-material-breakdown}, \textit{Unlikely} responses for the material items were quite rare. Across both few-shot and zero-shot data combined (a total of 1,867 cases), there was only one instance where all three workers responded with \textit{Unlikely}. Therefore, while these distributions provide insights into the reasoning behind such responses, they represent a minor portion of the overall precision evaluation of the material items.

\subsection{Breakdown of intrinsic evaluation results}
\label{sec:intrinsic-evaluation-breakdown}
To ensure that the evaluation on distinctive feature significance focuses on reliable and relevant part-subtypes, we exclude any part-subtypes that marked as \textit{Unlikely} during the precision evaluation. Additionally, part-subtypes were excluded if two out of three evaluators responded with \textit{Unable to answer} or \textit{Uncertain}. The final counts are 790 and 827 sets of one part and two subtypes, as shown in the table below.

\begin{table}[!htbp]
  \centering
    \begin{tabular}{lcc}
    \hline
    \textbf{Answer choices} & \textbf{Few-shot} & \textbf{Zero-shot}\\
    \hline
    Likely (3) & 110 & 60\\
    Likely (2) \& Unlikely (1) & 53 & 62\\
    Likely (2) \& Unable to answer (1) & 53 & 25\\
    Likely (2) \& Uncertain (1) & 6 & 5\\
    Likely (1) \& Unlikely (2) & 74 & 80\\
    Likely (1) \& Unable to answer (2) & 39 & 27\\
    Likely (1) \& Uncertain (2) & 0 & 0\\
    Likely (1) \& Unlikely (1) \& Unable to answer (1) & 58 & 35\\
    Likely (1) \& Unlikely (1) \& Uncertain(1) & 7 & 2\\
    Likely (1) \& Unable to answer (1) \& Uncertain(1) & 7 & 4\\
    Unlikely (1) \& Unable to answer (2) & 46 & 40\\
    Unlikely (1) \& Uncertain (2) & 1 & 1\\
    Unlikely (1) \& Unable to answer (1) \& Uncertain(1) & 7 & 10\\
    Unlikely (2) \& Unable to answer (1) & 71 & 103\\
    Unlikely (2) \& Uncertain (1) & 10 & 8\\
    Unlikely (3) & 235 & 325\\
    Unable to answer (1) \& Uncertain (2) & 0 & 0\\
    Unable to answer (2) \& Uncertain (1) & 8 & 6\\
    Unable to answer (3) & 5 & 34\\
    Uncertain (3) & 0 & 0\\
    \hline
    Total & 790 & 827\\
  \end{tabular}
  \caption{A detailed breakdown of responses on whether a specific part serves as \textbf{a main distinguishing factor} between two subtypes. The table shows the number of workers (indicated in parentheses) who responded in each category---\textit{Likely}, \textit{Unlikely}, \textit{Unable to answer}, and \textit{Uncertain}---as well as the number of questions asked in each case (displayed under the headers Few-shot and Zero-shot).}
\end{table}

\begin{table*}
  \centering
    \begin{tabular}{clcccc}
    \hline
    \textbf{Our data} & \textbf{Answer choices} & \textbf{Subtypes} & \textbf{Parts} & \textbf{Materials}\\
    \hline
    & Likely (3) & 316 & 650 & 641\\
    & Likely (2) \& Unlikely (1) & 13 & 55 & 67\\
    & Likely (2) \& Unable to answer (1) & 40 & 103 & 99\\
    & Likely (2) \& Uncertain (1) & 5 & 0 & 0\\
    & Likely (1) \& Unlikely (2) & 2 & 19 & 6\\
    & Likely (1) \& Unable to answer (2) & 25 & 60 & 27\\
    & Likely (1) \& Uncertain (2) & 0 & 0 & 0\\
    & Likely (1) \& Unlikely (1) \& Unable to answer (1) & 1 & 6 & 8\\
    Few-shot & Likely (1) \& Unlikely (1) \& Uncertain(1) & 1 & 1 & 0\\
    & Likely (1) \& Unable to answer (1) \& Uncertain(1) & 3 & 1 & 0\\
    & Unlikely (1) \& Unable to answer (2) & 3 & 1 & 2\\
    & Unlikely (1) \& Uncertain (2) & 0 & 0 & 0\\
    & Unlikely (1) \& Unable to answer (1) \& Uncertain(1) & 0 & 0 & 0\\
    & Unlikely (2) \& Unable to answer (1) & 1 & 4 & 0\\
    & Unlikely (2) \& Uncertain (1) & 0 & 0 & 0\\
    & Unlikely (3) & 1 & 21 & 0\\
    & Unable to answer (1) \& Uncertain (2) & 0 & 0 & 0\\
    & Unable to answer (2) \& Uncertain (1) & 3 & 1 & 0\\
    & Unable to answer (3) & 16 & 3 & 17\\
    & Uncertain (3) & 0 & 0 & 0\\
    Total & & 430 & 925 & 867\\
    \hline
    & Likely (3) & 1,489 & 608 & 759\\
    & Likely (2) \& Unlikely (1) & 76 & 92 & 75\\
    & Likely (2) \& Unable to answer (1) & 105 & 155 & 88\\
    & Likely (2) \& Uncertain (1) & 18 & 0 & 0\\
    & Likely (1) \& Unlikely (2) & 19 & 27 & 5\\
    & Likely (1) \& Unable to answer (2) & 33 & 41 & 41\\
    & Likely (1) \& Uncertain (2) & 0 & 0 & 0\\
    & Likely (1) \& Unlikely (1) \& Unable to answer (1) & 20 & 9 & 14\\
    Zero-shot  & Likely (1) \& Unlikely (1) \& Uncertain(1) & 3 & 0 & 0\\
    & Likely (1) \& Unable to answer (1) \& Uncertain(1) & 0 & 2 & 0\\
    & Unlikely (1) \& Unable to answer (2) & 9 & 6 & 2\\
    & Unlikely (1) \& Uncertain (2) & 0 & 0 & 0\\
    & Unlikely (1) \& Unable to answer (1) \& Uncertain(1) & 1 & 1 & 0\\
    & Unlikely (2) \& Unable to answer (1) & 7 & 9 & 0\\
    & Unlikely (2) \& Uncertain (1) & 0 & 1 & 1\\
    & Unlikely (3) & 6 & 33 & 1\\
    & Unable to answer (1) \& Uncertain (2) & 0 & 0 & 0\\
    & Unable to answer (2) \& Uncertain (1) & 0 & 4 & 0\\
    & Unable to answer (3) & 11 & 12 & 14\\
    & Uncertain (3) & 0 & 0 & 0\\
    Total & & 1,797 & 1,000 & 1,000\\
  \end{tabular}
  \caption{A detailed breakdown of the precision responses, showing the number of workers (indicated in parentheses) who responded in each category---\textit{Likely}, \textit{Unlikely}, \textit{Unable to answer}, and \textit{Uncertain}---along with the number of questions asked in each case (displayed under the headers Subtypes, Parts, and Materials.}
  \label{table:internal-precision-breakdown}
\end{table*}

\begin{table*}
  \centering
    \begin{tabular}{clcccc}
    \hline
    \textbf{Our data} & \textbf{Answer choices} & \textbf{Subtypes} & \textbf{Parts} & \textbf{Materials}\\
    \hline
    & Likely (3) & 8 & 98 & 337\\
    & Likely (2) \& Unlikely (1) & 22 & 125 & 237\\
    & Likely (2) \& Unable to answer (1) & 15 & 37 & 114\\
    & Likely (2) \& Uncertain (1) & 0 & 0 & 1\\
    & Likely (1) \& Unlikely (2) & 15 & 54 & 105\\
    & Likely (1) \& Unable to answer (2) & 0 & 23 & 13\\
    & Likely (1) \& Uncertain (2) & 0 & 0 & 0\\
    & Likely (1) \& Unlikely (1) \& Unable to answer (1) & 6 & 34 & 41\\
    Few-shot & Likely (1) \& Unlikely (1) \& Uncertain(1) & 0 & 0 & 0\\
    & Likely (1) \& Unable to answer (1) \& Uncertain(1) & 0 & 1 & 0\\
    & Unlikely (1) \& Unable to answer (2) & 0 & 5 & 4\\
    & Unlikely (1) \& Uncertain (2) & 0 & 0 & 0\\
    & Unlikely (1) \& Unable to answer (1) \& Uncertain(1) & 0 & 0 & 1\\
    & Unlikely (2) \& Unable to answer (1) & 1 & 8 & 3\\
    & Unlikely (2) \& Uncertain (1) & 0 & 0 & 0\\
    & Unlikely (3) & 5 & 23 & 11\\
    & Unable to answer (1) \& Uncertain (2) & 0 & 0 & 0\\
    & Unable to answer (2) \& Uncertain (1) & 0 & 0 & 0\\
    & Unable to answer (3) & 0 & 4 & 0\\
    & Uncertain (3)& 0 & 0 & 0\\
    Total & & 72 & 412 & 867\\
    \hline
    & Likely (3) & 23 & 486 & 417\\
    & Likely (2) \& Unlikely (1) & 22 & 231 & 291\\
    & Likely (2) \& Unable to answer (1) & 6 & 135 & 132\\
    & Likely (2) \& Uncertain (1) & 0 & 0 & 0\\
    & Likely (1) \& Unlikely (2) & 12 & 41 & 76\\
    & Likely (1) \& Unable to answer (2) & 0 & 52 & 40\\
    & Likely (1) \& Uncertain (2) & 0 & 0 & 0\\
    & Likely (1) \& Unlikely (1) \& Unable to answer (1) & 5 & 30 & 29\\
    Zero-shot  & Likely (1) \& Unlikely (1) \& Uncertain(1) & 0 & 0 & 1\\
    & Likely (1) \& Unable to answer (1) \& Uncertain(1) & 0 & 0 & 1\\
    & Unlikely (1) \& Unable to answer (2) & 0 & 4 & 4\\
    & Unlikely (1) \& Uncertain (2) & 0 & 0 & 0\\
    & Unlikely (1) \& Unable to answer (1) \& Uncertain(1) & 0 & 0 & 0\\
    & Unlikely (2) \& Unable to answer (1) & 0 & 1 & 5\\
    & Unlikely (2) \& Uncertain (1) & 0 & 0 & 0\\
    & Unlikely (3) & 13 & 6 & 4\\
    & Unable to answer (1) \& Uncertain (2) & 0 & 0 & 0\\
    & Unable to answer (2) \& Uncertain (1) & 0 & 0 & 0\\
    & Unable to answer (3) & 0 & 14 & 0\\
    & Uncertain (3) & 0 & 0 & 0\\
    Total & & 81 & 1,000 & 1,000\\
  \end{tabular}
  \caption{A detailed breakdown of the recall responses, showing the number of workers (indicated in parentheses) who responded in each category---\textit{Likely}, \textit{Unlikely}, \textit{Unable to answer}, and \textit{Uncertain}---along with the number of questions asked in each case (displayed under the headers Subtypes, Parts, and Materials.}
\end{table*}

\newpage

\subsection{Breakdown of dataset-comparison evaluation results}
\label{sec:dataset-comparison-breakdown}
\begin{table*}[!htbp]
  \centering
  \begin{tabular}{l@{\hskip-0.5mm}clllccc}
    \hline
    \multirow{2}{*}{\textbf{Category}} & \multirow{2}{*}{\textbf{Response}} & & \multirow{2}{*}{\textbf{Score}} & \multicolumn{3}{c}{\textbf{Number of responses}}\\
    & & & & \textbf{Few} & \textbf{Zero} & \textbf{Human}\\
    \hline
    \multirow{19}{*}{Subtype} & \multirow{5}{*}{Familiarity} & with most subtypes & +1.0 & 56 & 73 & 30\\
    & & with some subtypes & +0.5 & 12 & 21 & 10\\
    & & with few subtypes & -0.5 & 13 & 21 & 4\\
    & & with none & -1.0 & 3 & 2 & 2\\
    & & N/A, no subtypes listed & 0.0 & 96 & 63 & 134\\
    \cmidrule{2-7}
    & \multirow{4}{*}{Coverage} & comprehensive & +1.0 & 67 & 103 & 34\\
    & & missing one & -0.5 & 7 & 1 & 2\\
    & & missing two or more & -1.0 & 17 & 9 & 21\\
    & & unfamiliar or uncertain & 0.0 & 4 & 9 & 4\\
    \cmidrule{2-7}
    & \multirow{4}{*}{Level of detail} & appropriate & +1.0 & 78 & 99 & 42\\
    & & overly specific & -1.0 & 3 & 17 & 3\\
    & & too general or broad  & -1.0 & 0 & 0 & 0\\
    & & unfamiliar or uncertain & 0.0 & 3 & 1 & 1\\
    \cmidrule{2-7}
    & \multirow{3}{*}{Clarity \& distinction} & well-categorized & +1.0 & 79 & 102 & 45\\
    & & some subtypes overlap & -1.0 & 2 & 14 & 0\\
    & & unfamiliar or uncertain & 0.0 & 3 & 1 & 1\\
    \cmidrule{2-7}
    & Consistency in & consistent & +1.0 & 80 & 113 & 45\\
    & level of detail or style & inconsistent & -1.0 & 2 & 3 & 0\\
    & & unfamiliar or uncertain & 0.0 & 2 & 1 & 1\\
    \hline
    \multirow{16}{*}{Part} & \multirow{4}{*}{Focus on essential parts} & all essential parts & +1.0 & 167 & 163 & 139\\
    & & some unnecessary parts & -1.0 & 5 & 12 & 2\\
    & & unfamiliar or uncertain & 0.0 & 1 & 1 & 4\\
    & & N/A, no parts listed & 0.0 & 7 & 4 & 35\\
    \cmidrule{2-7}
    & \multirow{4}{*}{Level of detail} & appropriate & +1.0 & 165 & 171 & 131\\
    & & overly detailed & -1.0 & 1 & 4 & 1\\
    & & too general or broad  & -1.0 & 6 & 0 & 11\\
    & & unfamiliar or uncertain & 0.0 & 1 & 1 & 2\\
    \cmidrule{2-7} 
    & \multirow{5}{*}{Clarity \& distinction} & well-identified & +1.0 & 169 & 170 & 140\\
    & & correctly no parts listed & +1.0 & 5 & 1 & 16\\
    & & incorrectly no parts listed & -1.0 & 2 & 3 & 21\\
    & & some parts overlap & -1.0 & 3 & 4 & 1\\
    & & unfamiliar or uncertain & 0.0 & 1 & 2 & 2\\
    \cmidrule{2-7} 
    & \multirow{3}{*}{Consistent level of detail} & consistent & +1.0 & 172 & 170 & 141\\
    & & inconsistent & -1.0 & 0 & 5 & 3\\
    & & unfamiliar or uncertain & 0.0 & 1 & 1 & 1\\
    \hline
    \multirow{3}{*}{Material} & \multirow{3}{*}{Clarity \& distinction} & well-identified & +1.0 & 165 & 154 & 171\\
    & & some materials overlap & -1.0 & 13 & 23 & 4\\
    & & unfamiliar or uncertain & 0.0 & 2 & 3 & 5\\
    \bottomrule
    \multicolumn{4}{r}{\textbf{Aggregate score}} & 1,142 & 1,222.5 & 867\\
    \multicolumn{4}{r}{\textbf{Number of rated responses}} & 1,292 & 1,458 & 1,019\\
    \multicolumn{4}{r}{\textbf{Mean score}} & 88.39 & 83.85 & 85.08\\
  \end{tabular}
  \caption{Comparison of few-shot, zero-shot, and human-annotated datasets across multiple response categories, showing the number of responses and scores based on rating criteria. The table concludes with each dataset's mean score, calculated as the ratio of the aggregate score to the total number of rated (non-neutral) responses.}
  \label{table:dataset-comparison-breakdown}
\end{table*}

\newpage

\section{Parts and materials annotation criteria}
\label{sec:human-annotation-criteria}
1) Check the \textit{Non-physical entity} column if \textit{Item} under the first column is a non-physical entity or is a physical entity with independently functioning parts (e.g., salad bar).
\newline\newline
2) Specify the subtypes in the \textit{Subtype} column if some of the subtypes are made of different materials or their part structure is different from one another.
 \newline\newline
3) If the Wikipedia article specifies a subtype name, please use it possibly without modification. If there are more than one name mentioned for the same subtype and one of them has a Wikipedia article linked to it, please use the linked article's title name. Title names should be case-sensitive and keep any content within parentheses (otherwise, my code won't be able to find the right article). As long as we keep the name correctly, we don't need any special notations for subtypes that exist as a Wikipedia article.
\newline\newline 
4) For subtypes that aren't mentioned in the articles, try to use pre-noun modifiers in the subtype names like `lace-less' or `zippered' instead of prepositions added to the noun.
 \newline\newline
5) We allow subsubtypes of an item (i.e., when subtypes have their own subtypes) only when the subtype has its own Wikipedia article. We put the item in the first column and its subtype in the \textit{Subtype} column, and create new rows to put the subtype in first column and subsubtypes in the \textit{Subtype} column. For the subtypes that have their own subtypes (i.e., subsubtypes of the main item), leave out any annotations for the subtypes, but fill in the rows for subsubtypes. Please keep the names consistently and again, please do not alter Wikipedia title names.
 \newline\newline
6) Put asterisks around the parts that are not found in the Wikipedia articles.
\newline\newline
7) Check the \textit{Optional} column for any optional parts (e.g., sprinkler attached to the spout of a watering can or a lid on a paper cup).
\newline\newline
8) We need to make a distinction among the three types of materials annotated:\newline
\hspace*{0.4cm} A. ``Proper materials'': Materials that are found in a proper sentence. Any of the following\\\hspace*{4.0cm}information mentioned in the object's or the subtype's Wikipedia article\\\hspace*{4.0cm}satisfy \textit{M} to be ``Proper materials'' for a subtype \textit{S}, an object \textit{O} (a supertype of\\\hspace*{4.0cm}\textit{S}), and parts \textit{P1} and \textit{P2}. Sentences in past tense are considered as proper\\\hspace*{4.0cm}sentences as well.\\
\hspace*{1.0cm}- \textit{O} is made of \textit{M} (e.g., ``a cup is made of glass'' or ``a glass cup'')\\
\hspace*{1.0cm}- \textit{S} is made of \textit{M}\\
\hspace*{1.0cm}- \textit{P1} of \textit{S} or \textit{O} is made of \textit{M}\\
\hspace*{1.0cm}- \textit{P2} of \textit{S} or \textit{O} is made of \textit{M}\\
\hspace*{1.0cm}We allow \textit{M} to be the proper materials for \textit{P1} even when the article mentions \textit{P2} is made of \textit{M};\\\hspace*{1.0cm}this is to enable \textit{uniform materials} annotations in 11) below.\newline
\hspace*{0.4cm} B. ``Found materials'': Materials, other than what's found in A, that are found somewhere in the\\\hspace*{4.0cm}Wikipedia article.\\
\hspace*{0.4cm} C. ``Inferred materials'': Materials that are gathered from inference or googling.\newline
We won't put any conjunctions under these columns, but rather we will put it under the \textit{Conjunction} column. If there is only one material, put ``-'' instead for a conjunction. Annotating this way, we won't need any asterisks for materials. Just commas.
\newline\newline
9) We're allowed to use * entity * under \textit{Predicate} to indicate that the part consists of further parts.
\newline\newline
10) We're allowed to use * internal mechanism * as a part.
\newline\newline
11) For two or more major parts of an object (e.g., a jacket's body, collar, sleeves \& pockets) that are made uniformly of one or alternative materials, please put one of the main parts as the \textit{Part 1} column, check the \textit{Uniform materials} column, and for any other major parts that would have the same materials as listed in the \textit{Proper materials}, \textit{Found materials}, and \textit{Interred materials} columns, you may leave the material cells empty. For uniform materials to apply, the set of materials under the proper, found, and inferred materials should be repeated exactly for the proper parts. For example, let's say for a soft pencil case, we have \textit{Uniform materials} checked and\\
\hspace*{1.0cm}a. Part1: ``pouch'' -- plastic, leather, cotton\\
\hspace*{1.0cm}b. Part2: ``zipper pull tab'' -- [empty]\\
\hspace*{1.0cm}c. Part3: ``zipper'' – metal\\
this annotation means that the zipper pull tab is made of whatever materials that a pouch is made of. If the zipper pull tab can be also made of metal, then the example should be rather,\\
\hspace*{1.0cm}b. ``zipper pull tab'' made of plastic, leather, cotton, metal\\
and \textit{Uniform materials} shouldn't be checked.\\
Please use the \textit{Uniform materials} column conservatively so that it is not over-used for non-uniform materials. For instance, if a travel mug's body is made of plastic or stainless steel and its lid is made of plastic or stainless steel, but if it's common for a travel mug to have a stainless-steel body and a plastic lid, then \textit{Uniform materials} shouldn't be checked.
\newline\newline
12) Parts and materials can be repeated across different rows. We want to get all the part/material information for a type/subtype just from looking at each row.
\newline\newline
13) Use ``[A] and/or [B]'' to indicate materials for an object that satisfies all of the following:\\
\hspace*{1.0cm}- entirely made of A\\
\hspace*{1.0cm}- entirely made of B\\
\hspace*{1.0cm}- possibly made of a combination of A and B\\

\end{document}